%% file: New_Main.tex
\pdfoutput=1

\documentclass[11pt]{article}

\usepackage{acl}
\usepackage{fdsymbol}

\usepackage{times}
\usepackage{latexsym}
\usepackage{makecell}
\usepackage{tablefootnote}
\usepackage[toc,page]{appendix}
\usepackage{pifont}
\usepackage[most]{tcolorbox}
\tcbuselibrary{listings,breakable,skins}

\usepackage[most]{tcolorbox}
\tcbuselibrary{listings,breakable,skins}
\usepackage{listings}
\lstdefinestyle{promptstyle}{
  basicstyle=\ttfamily\scriptsize,
  breaklines=true,
  breakatwhitespace=true,
  columns=fullflexible,
  keepspaces=true,
  showstringspaces=false,
  tabsize=2
}
\usepackage{algorithm2e}
\usepackage{algpseudocode}
\usepackage{enumitem}
\usepackage{fancyvrb}
\usepackage{cuted, tcolorbox}
\usepackage[T1]{fontenc}

\usepackage{booktabs}
\usepackage{makecell}
\usepackage[table]{xcolor}
\usepackage{graphicx}
\usepackage{fontawesome5}

\usepackage[utf8]{inputenc}

\usepackage{microtype}

\usepackage{hyperref}
\usepackage{booktabs}
\usepackage{graphicx}
\usepackage{makecell}
\graphicspath{{./imgs/}}
\usepackage{footmisc}
\usepackage{alltt}
\usepackage{floatrow}
\usepackage{arydshln}
\usepackage{microtype}

\usepackage{caption}
\usepackage{subcaption}
\usepackage{array, makecell} %
\usepackage{amsmath}
\usepackage{pifont}
\usepackage{tabularx,colortbl}

\usepackage{tikz}
\usepackage{collcell}

\usepackage{etoolbox}

\newcolumntype{?}{!{\vrule width 1.5pt}}

\newtoggle{inTableHeader}
\toggletrue{inTableHeader}
\newcommand*{\StartTableHeader}{\global\toggletrue{inTableHeader}}%
\let\OldTabular\tabular%
\let\OldEndTabular\endtabular%
\renewenvironment{tabular}{\StartTableHeader\OldTabular}{\OldEndTabular\StartTableHeader}%

\newcommand*{\MinNumber}{-1.0}%
\newcommand*{\MidNumber}{0.0} %
\newcommand*{\MaxNumber}{1.0}%

\newcommand{\ApplyGradient}[1]{%
  \iftoggle{inTableHeader}{#1}{
    \ifdim #1 pt > \MidNumber pt
        \pgfmathsetmacro{\PercentColor}{max(min(100.0*(#1 - \MidNumber)/(\MaxNumber-\MidNumber),100.0),0.00)} %
        \hspace{-0.33em}\colorbox{yellow!\PercentColor!blue}{#1}
    \else
        \pgfmathsetmacro{\PercentColor}{max(min(100.0*(\MidNumber - #1)/(\MidNumber-\MinNumber),100.0),0.00)} %
        \hspace{-0.33em}\colorbox{blue!\PercentColor!blue}{#1}
    \fi
  }}
\newcolumntype{R}{>{\collectcell\ApplyGradient}c<{\endcollectcell}}

\input{math-com}

\usepackage[nameinlink]{cleveref}
\crefformat{section}{\S#2#1#3} 
\crefname{algorithm}{Alg.}{Algs.}
\crefformat{subsection}{\S#2#1#3}
\Crefname{equation}{Eq.}{Eqs.}
\Crefname{figure}{Fig.}{Figs.}

\usepackage[colorinlistoftodos,prependcaption,textsize=tiny]{todonotes}

\usepackage{soul}

\usepackage{float}
\usepackage{makecell}
\definecolor{azure}{rgb}{0.0, 0.5, 1.0}
\definecolor{darkbrown}{rgb}{0.4, 0.26, 0.13}
\definecolor{moonstoneblue}{rgb}{0.45, 0.66, 0.76}
\usepackage{multirow}
\usepackage{multicol}
\usepackage{hhline}
\usepackage[export]{adjustbox}

\usepackage[table]{xcolor}

\newcommand{\model}{{\textsc{\textbf{DataReel}}}}

\definecolor{datareelbench1}{RGB}{0,0,0}        
\definecolor{datareelbench2}{RGB}{230,126,34}  
\newcommand{\datareelbench}{\textsc{\textcolor{datareelbench1}{Data}\textcolor{datareelbench2}{Reel}}}

\title{
\raisebox{0.1em}{
\begin{adjustbox}{valign=c}
    \includegraphics[height=1.0em]{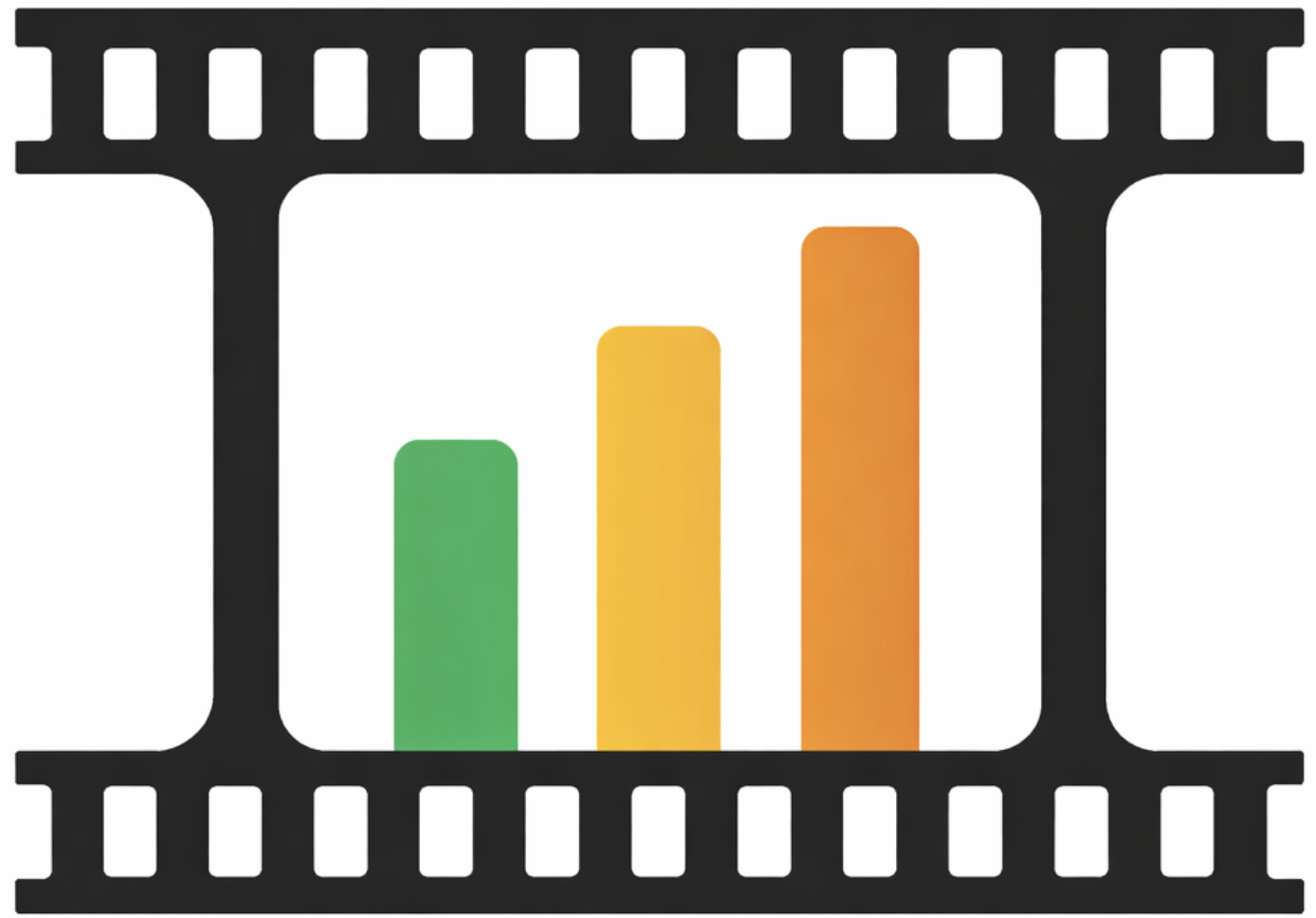}
\end{adjustbox}}
\hspace{-0.6em}
\textsc{\textbf{
\textcolor{datareelbench1}{Data}\textcolor{datareelbench2}{Reel}
}}\hspace{-.25em}: Automated Data-Driven Video Story Generation \\with Animations
}

\author{
Ridwan Mahbub$^{\spadesuit}$, \
Syem Aziz$^{\triangle}$, \ Mizanur Rahman$^{\spadesuit}$, \ Mahir Ahmed$^{\spadesuit}$, \\
\bf  \ Shadikur Rahman$^{\spadesuit}$ \
Shafiq Joty$^{\clubsuit\diamondsuit}$, \ Enamul Hoque$^{\spadesuit}$ \\
$^\spadesuit$York University, 
$^\triangle$Bangladesh University of Business and Technology, \\
$^\clubsuit$Nanyang Technological University 
$^\diamondsuit$Salesforce AI Research \\
\{rmahbub, mizanurr, mahmed, shadikur, enamulh\}@yorku.ca\\ syemaziz@bubt.edu.bd, sjoty@salesforce.com
}

\begin{document}
\maketitle

\begin{abstract}
Data videos combine animated visualizations with synchronized narration to communicate quantitative information and are widely used in journalism, education, and public communication.  Automatically generating them requires deciding what story to tell, designing effective visualizations, and producing executable animations synchronized with narration. Despite rapid progress in vision-language models (VLMs), it remains unclear how well they can perform this task from a high-level communicative intent, largely because no standardized benchmark exists. We introduce \datareelbench{}, a
benchmark for automated data-driven video story generation containing 328 real-world
\emph{data reels}. Given a data table, a communicative
intent, a target duration, and a style reference image, a model must generate
executable animation code with  synchronized subtitles, which we render and evaluate.
Evaluating eight proprietary and open-weight VLMs reveals a substantial capability gap: open-weight models exhibit execution failure rates of up to 39.8\%, whereas proprietary models achieve stronger overall performance. Yet even the best-performing  models frequently generate static charts, subtitle–animation desynchronization, unstable layouts, and poor adherence to the reference style. We further introduce a strong agentic baseline that decomposes generation into planning, coding, and verification, consistently outperforming direct prompting in both human and automatic evaluations. The task remains far from solved; we release \datareelbench{} at \url{https://github.com/vis-nlp/DataReel} to support future work.
\end{abstract}

\section{Introduction}
\vspace{-1mm}

\begin{figure}[t!]
     \centering
        \includegraphics[width=\textwidth]{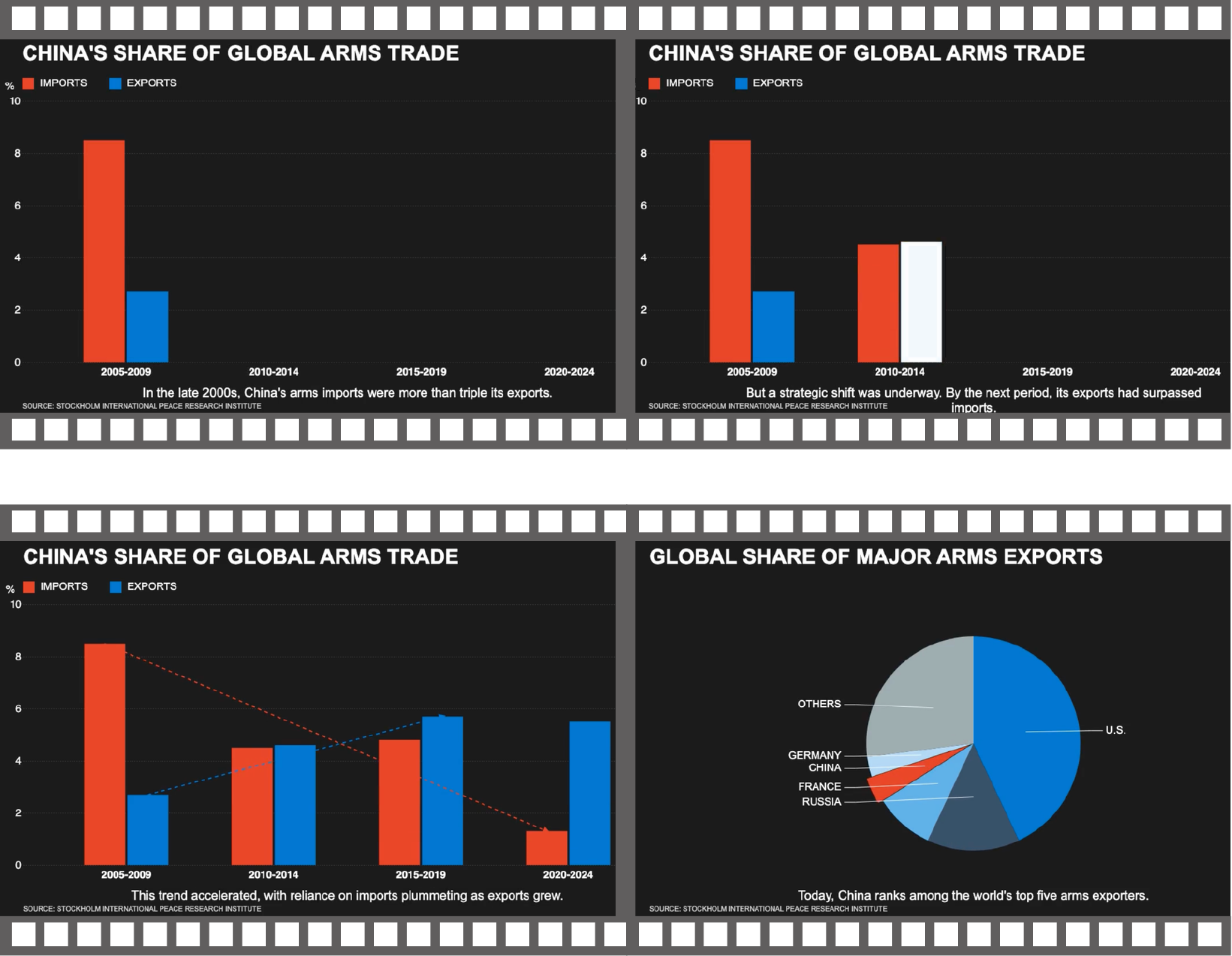}
         \caption{
         A \emph{data reel} generated by the multi-agent baseline using Gemini 2.5 Pro. The video version is available \href{https://www.youtube.com/watch?v=hDj_u2Me6bk}{online}. The clip illustrates China’s rise in the global arms industry, showing its transition from a net arms importer to one of the world’s top five arms exporters. The animation scenes are accompanied by subtitles at the bottom. The actual video is extracted from the YouTube channel Wall Street Journal\footnotemark
          }
    \label{fig:example_gpt4o_mini}
    \vspace{-3mm}
\end{figure}

\footnotetext{\url{https://youtu.be/M6CNlvXkR7k?si=05OMIyLBj2PdeQmv&t=97}}

Visual data stories have become an effective medium for data communication, combining visualizations with coherent narratives~\citep{hullman2011visualization}. Although static visual data stories remain common, the rapid proliferation of video-sharing platforms has driven growing interest in data videos, commonly referred to as data videos~\citep{amini2015understanding}. 
Widely adopted across domains, particularly digital
journalism, data videos are generally more accessible and easier to follow than static
formats, and often achieve higher user engagement through animation, narration, and
temporal sequencing~\citep{amini2018hooked}.

Structurally, a typical data video 
consists of multiple short, chart-centric animation clips, often accompanied by voiceover narration and interleaved with real-world images or footage~\citep{cheng2022investigating}. With the rise of short-form videos and reels on social media platforms~\citep{roberts2025technology}, such chart-centric animated clips have become increasingly common. In this paper, we refer to these clips as \emph{data reels}, which can function either as standalone videos or as components within longer videos.

~\Cref{fig:example_gpt4o_mini} shows an automatically generated  data reel, illustrating China’s share in global arms trade. In the first scene, arrow animations emphasize the decline in China’s arms imports, highlighting their lowest level during 2020--2024. Later, animated blue bars illustrate rising arms exports, before the chart transitions into a pie chart emphasizing China’s position among top exporters. 

Despite their impact, creating data reels remains challenging. Unlike static chart design~\cite{rahman2025text2vis} or traditional video creation~\cite{amini2016authoring}, data reels require coordinated visual encoding, temporal progression, and narration, along with expertise in visualization design, animation, and video-editing tools. Combining data interpretation, narrative planning, animation design, and audiovisual alignment makes the task difficult even for humans. In practice, authors often rely on low-level animation and video-editing tools, resulting in substantial manual effort and a high barrier to entry, particularly for non-experts~\citep{wang2024wonderflow}. The challenge is further amplified by the need for tight animation--narration alignment, where even minor inconsistencies can disrupt narrative clarity and viewer attention.

Prior research on data-driven storytelling has primarily focused on static visual stories, introducing narrative design spaces, visual encodings, and authoring tools to support effective communication~\citep{segel2010narrative,hullman2013adeeper,lan2022kinecharts,mckenna2017visualnarrative,shi2021communicating,shi2021understanding}. While these studies provide valuable guidance, manually creating high-quality data stories remains time-consuming and cognitively demanding. Automated approaches have also been explored~\citep{shi2019taskoriented,shi2021calliope,wang2020datashot}, but they typically generate isolated facts or simple annotations, falling short of producing coherent, engaging stories—let alone animated video narratives like \emph{data reels}.

Recent advances in large language models (LLMs) have spurred work on  data-centric tasks, including chart summarization~\citep{kantharaj-etal-2022-chart,rahman2023chartsumm}, chart question answering~\citep{masry-etal-2022-chartqa,kantharaj-etal-2022-opencqa}, and natural language story generation~\citep{zhou2023recurrentgpt,xie-riedl-2024-creating}. More recent work has explored LLMs for static data story
generation~\citep{islam-etal-2024-datanarrative,wang2025jupybara}, alongside automated
systems that produce data videos from an existing chart or
narrative texts~\citep{shen2023dataplayer,ying2024livechart,shao2025narrative}. However, whether general-purpose LLMs can generate complete \emph{data reels} from a data table and a short communicative intent remains underexplored, largely due to the lack of standardized benchmarks.

To address this gap, we introduce \datareelbench, a benchmark for automated data-driven video story generation comprising 328 real-world \emph{data reels} collected from  14 YouTube popular channels spanning journalism, education, and explanatory media. Given a data table, a communication intent, a target animation duration, and a reference image specifying the visual style, LLMs are required to generate code-based animated visualizations—using popular chart animation libraries such as D3.js—along with subtitles that remain coherent with the narrative. This task requires the model to jointly handle data interpretation, synchronized narrative generation, animation planning, and code generation, making the generation of \emph{data reels} a particularly challenging problem. Our experiments also indicate that even state-of-the-art models and agents, such as Claude Opus 4.5 and Gemini 2.5 Pro frequently produce unstable animations, incorrect layouts, and poor adherence to the reference style.

To establish a strong reference baseline for this task, we further introduce an agentic framework that decomposes \emph{data reel} generation into planning, generation, and verification stages. Motivated by recent successes of agentic LLM systems in complex planning tasks~\citep{yang2023autogpt,wang2023voyager,chen2024autoagents} and visualization code generation~\citep{islam-etal-2024-datanarrative,rahman2025text2vis}, the framework mirrors key aspects of the human authoring process while enabling more structured generation than direct prompting. Through automatic and human evaluations, we show that this multi-stage baseline consistently outperforms direct prompting and provides a strong foundation for future work on automated data video generation.

Our main contributions are: 
\emph{(i)} We introduce \datareelbench{}, the first benchmark for automated \emph{data reel} generation, comprising 328 real-world animated data stories paired with structured data, narration, and animation annotations;
\emph{(ii)} we introduce a multi-agent VLM framework as a baseline for generating animated data visualizations aligned with narration; and
\emph{(iii)}  we conduct extensive automatic and human evaluations that characterize the current capabilities and limitations of LLMs for data-driven video storytelling.

\begin{figure*}[t]
    \centering
\caption{Overview of our benchmarking process: (1) We construct \datareelbench{} by collecting and annotating real-world \emph{data reels} and extracting their underlying chart data. (2) Given data and communicative intent, along with a target animation duration and a reference image specifying the visual style, one or more agentic VLMs generate animation code. (3) The code is executed within a rendering framework to produce the final data video. (4) The generated videos are evaluated using both human assessment and a VLM judge.}   
     \label{fig-Methodology}
    \includegraphics[width=0.98\textwidth]{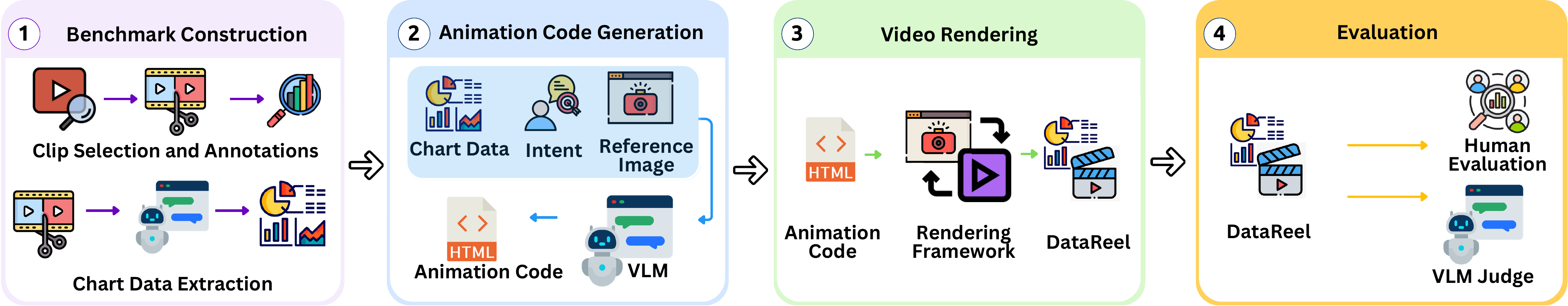} 
    \vspace{-2mm}
\end{figure*}

\vspace{-2mm}

\section{Related Work}
\label{sec:RelatedWorks}
\vspace{-1mm}
\textbf{Story Generation Tasks.} Automated story generation focuses on producing coherent sequences of events under open-ended constraints~\citep{li2013story}. Prior work has explored textual~\citep{kumar2006algorithms}, visual~\citep{li2019storygan,cohn2020visual}, and multimodal~\citep{bensaid2021fairytailor} story generation, often focusing on narrative coherence and character-centric progression.

In contrast, data-driven storytelling generates multimodal narratives in which visualizations communicate patterns, trends, and outliers while accompanying text explains and contextualizes them~\citep{riche2018data,segel2010narrative,hullman2013contextifier}. Early work focused on extracting and ranking salient insights from data tables~\citep{rui2019quickinsights,bo2017extracting}. Systems such as DataShot~\citep{wang2020datashot} and Calliope~\citep{shi2021calliope} pair visualizations with textual captions, while later tools including Erato~\citep{sun2023erato} and Socrates~\citep{wu2024socrates} incorporate user interaction to guide narrative structure and content selection. More recent work has explored LLMs for static data story generation~\citep{he2024leveraging,islam-etal-2024-datanarrative,wang2025jupybara}. 

Unlike static storytelling, data videos require synchronized coordination across multiple components, including data correctness, temporal sequencing, visual transitions, subtitle alignment, executable animation code, and stylistic consistency. Consequently, automated data-video generation is not a straightforward extension of static storytelling, but a distinct multimodal generation problem requiring coordinated narrative, visual, temporal, and programmatic elements.

\noindent\textbf{Automated Data Video Generation.}
Early systems relied on templates and heuristic rules~\citep{amini2016authoring,shi2021autoclips}, which limit expressiveness, while later tools support interactive
authoring~\citep{wang2024wonderflow}. More recent work has explored automatic data video generation from an existing chart or written narrative.  Data Player~\citep{shen2023dataplayer}
links an author-written narration to a static visualization and recommends animation
presets through constraint solving; Live Charts~\citep{ying2024livechart} parses a static
chart and generates component-level narration with matching animations; and Narrative
Player~\citep{shao2025narrative} turns a narrative paragraph and its data table into an optimized visualization sequence with transitions. All three are automatic once their inputs are provided, requiring no manual animation authoring. These systems assemble videos from predefined animation operations over visualizations that are supplied or system-selected, so their outputs render by construction. Our task is open-ended by comparison: a model must plan a narrative
from a one-sentence intent, choose the visualization design, and generate executable code that may fail to render.  They are also evaluated through example galleries, user studies, and expert interviews rather than shared benchmarks, as is recent work on LLM-based multi-agent generation~\citep{shen2024dataStory}, leaving model capabilities difficult to compare. \datareelbench{} defines this task and evaluates models on a common set of real-world reels.

\begin{figure*}[t]
    \centering
    \begin{subfigure}[t]{0.33\textwidth}
        \centering
        \includegraphics[width=\linewidth]{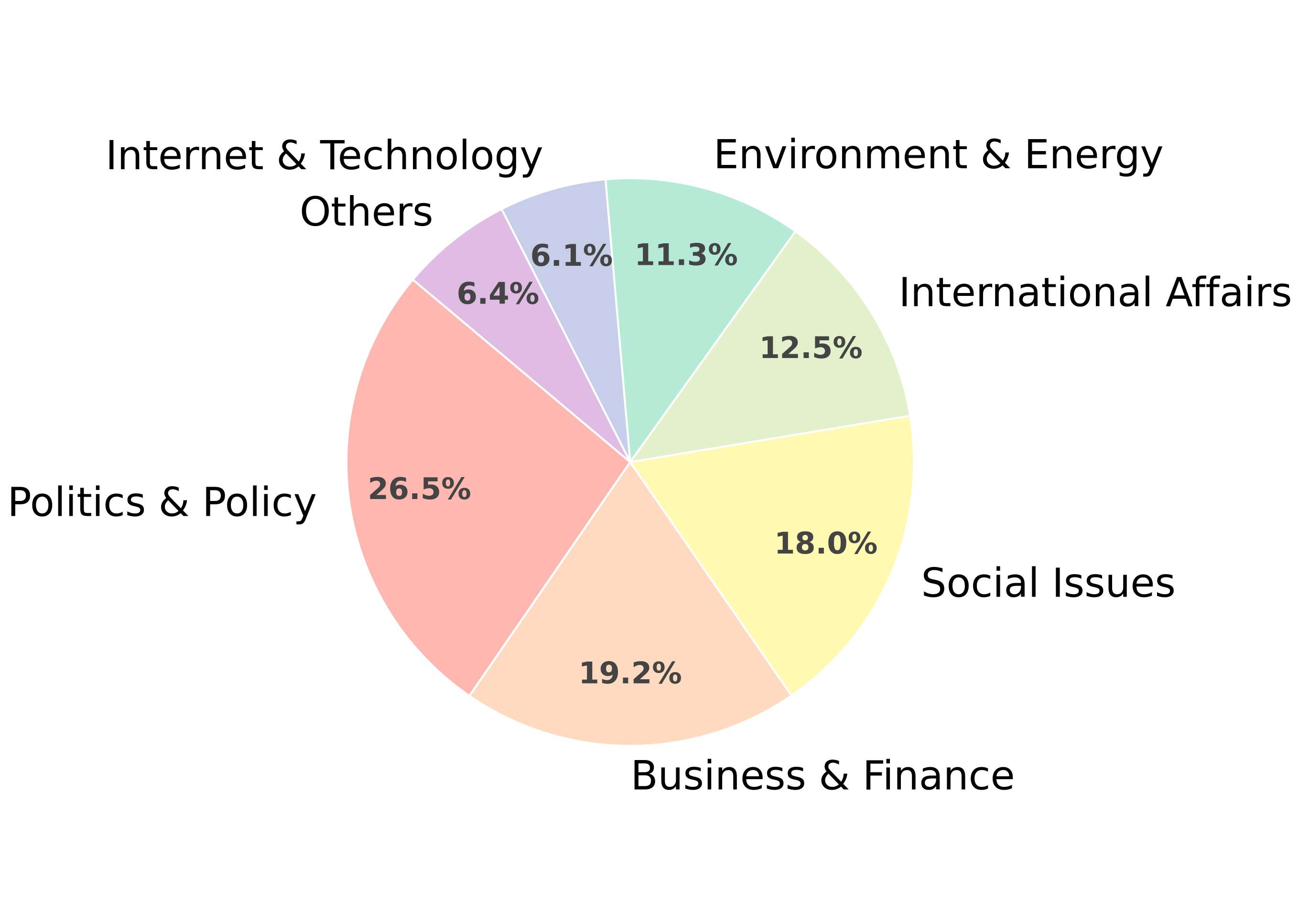}
        \caption{Topic distribution.}
        \label{fig:topic_distribution}
    \end{subfigure}
    \hfill
    \begin{subfigure}[t]{0.31\textwidth}
        \centering
        \includegraphics[width=\linewidth]{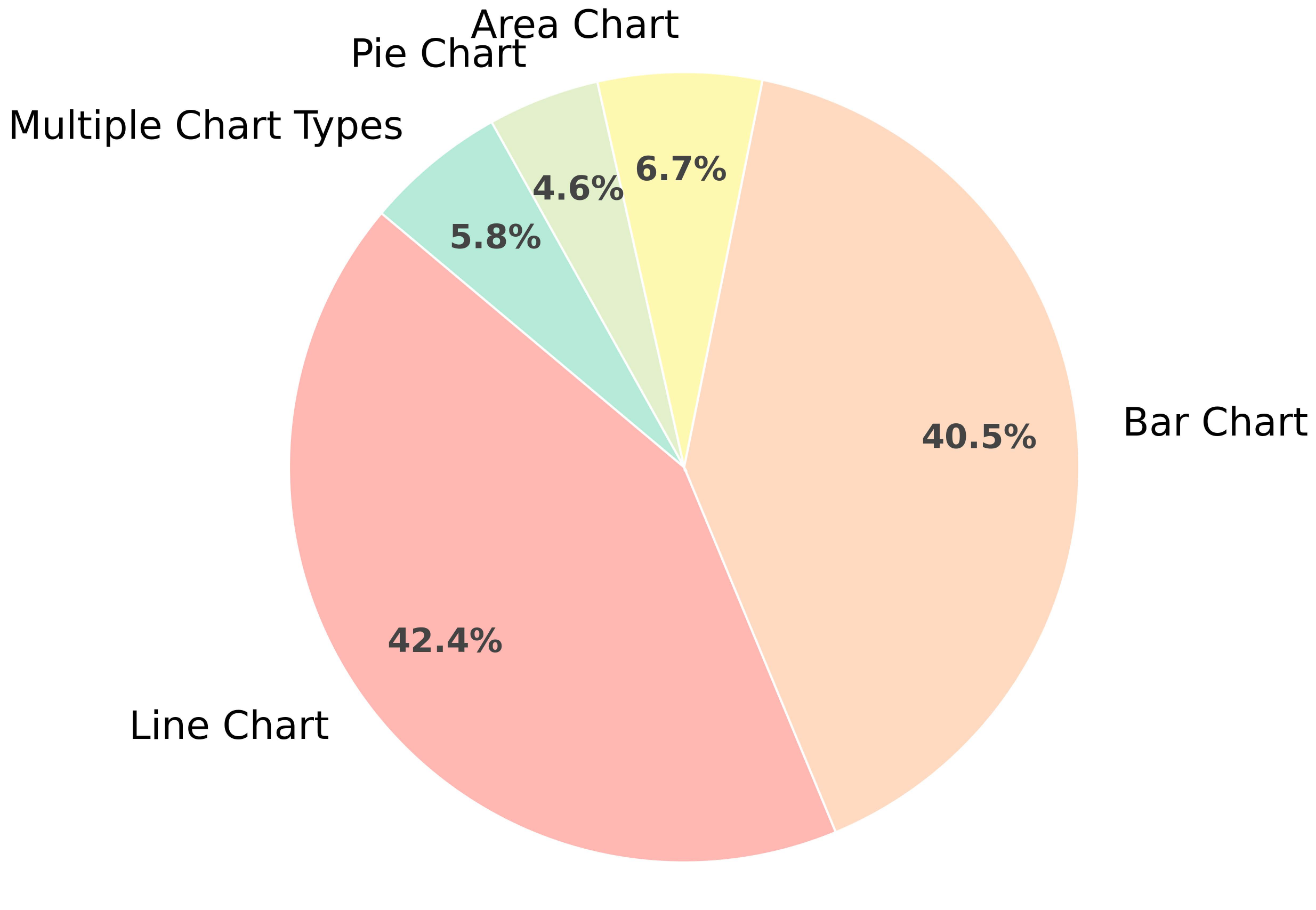}
        \caption{Chart type distribution.}
        \label{fig:chart_distribution}
    \end{subfigure}
    \hfill
    \begin{subfigure}[t]{0.32\textwidth}
        \centering
        \includegraphics[width=\linewidth]{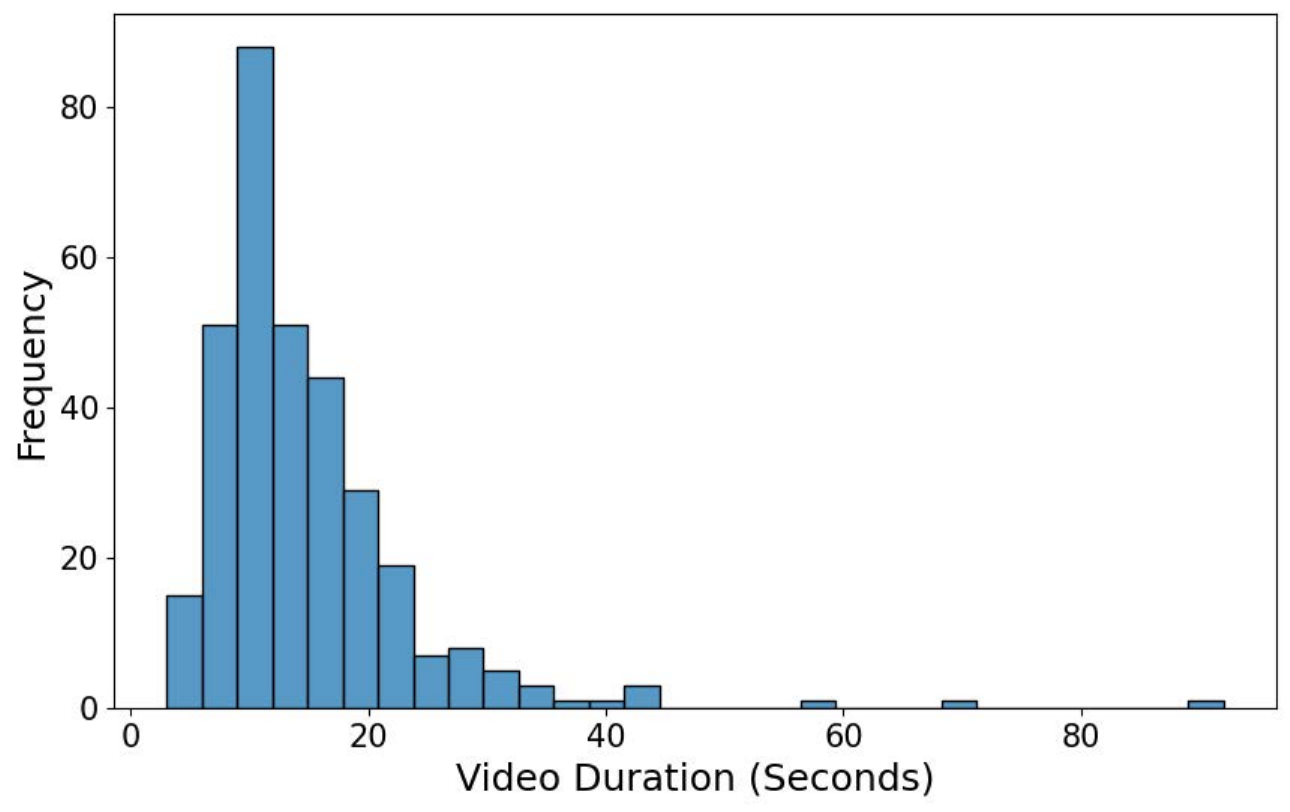}
        \caption{Video duration distribution.}
        \label{fig:durations}
           \vspace{-3mm}        
    \end{subfigure}
    \caption{Dataset characteristics of \datareelbench{}. Distribution of (a) topics, (b) chart types, and (c) \emph{data reel} durations.
    \vspace{-4mm}
    }
    \label{fig:dataset_overview}
    \vspace{-4mm}
\end{figure*}
\vspace{-2mm}

\noindent\textbf{Chart-related Downstream Tasks.}
Several downstream tasks associated with charts have been proposed in recent years. Chart question answering focuses on answering factual or reasoning-based questions about charts~\citep{masry-etal-2022-chartqa,masry-etal-2025-chartqapro}, while open-ended chart QA generates explanatory responses~\citep{kantharaj-etal-2022-opencqa}. Chart summarization aims to produce concise textual descriptions of charts~\citep{kantharaj-etal-2022-chart,tang2023vistext}, whereas chart-to-table tasks extract underlying data from chart images~\citep{choi2019visualizing,masry2023unichart,masry2024chartinstruct}. Others focus on verifying claims about charts~\citep{akhtar-etal-2023-reading, akhtar2024chartcheck}. Recent work has also focused on automated visualization generation from text queries~\citep{rahman2025text2vis}; however, they focused on static charts rather than videos. To our knowledge, there is no existing benchmark on data video generation.

\vspace{-2mm}
\section{Benchmark Data Construction}

\vspace{-2mm}

Our benchmark comprises \emph{data reels}—short, animated data video clips that communicate insights through chart-based visualizations synchronized with narration. Unlike conventional news videos, which primarily rely on real-world footage accompanied by spoken commentary, \emph{data reels} place animated data visualizations at the core of the narrative, making them a distinct and challenging medium for automated generation.

An overview of our benchmarking process is shown in \Cref{fig-Methodology}. We constructed \datareelbench{} through a four-stage pipeline: \textit{(i)} data video sourcing, \textit{(ii)} \emph{data reel} selection, \textit{(iii)} \emph{data reel} annotation, and \textit{(iv)} LLM-based data extraction. The benchmark construction pipeline (Step 1 in \Cref{fig-Methodology}) is illustrated in \Cref{fig:Benchmark}. Each data reel contains tightly coupled multimodal annotations, including structured data, narration, animation intent, temporal alignment, and executable generation targets, making collection substantially more expensive and time-consuming than static chart benchmarks.

\noindent \textbf{\textit{(i)} Data Video Sourcing:}
We focused on collecting \emph{data reels} from real-world, popular sources where data-driven storytelling is routinely practiced. In contrast to standalone data videos, \emph{data reels} often appear sparsely embedded within longer news or documentary videos, making their identification non-trivial.  To guide sourcing, we reviewed prior analyses of data videos in journalism~\citep{amini2015understanding,cheng2022investigating} and examined the channels used in those studies. We then augmented this list through an exploration of news-centric YouTube channels that regularly publish videos containing animated data visualizations. This process resulted in a curated set of 14 channels that consistently produce data-driven video content.

\noindent \textbf{\textit{(ii)} \emph{Data Reel} Selection:}
Each annotator was assigned a subset of channels identified in the sourcing stage. Annotators skimmed through videos to identify segments containing \emph{data reels} and recorded precise start and end timestamps for each clip. For every extracted clip, annotators also collected the corresponding narration transcript. They examined the channels in reverse chronological order to prioritize recent content, as animation quality and production practices have improved over time. A single video could yield multiple \emph{data reels} when distinct animated visualization segments were present.

\noindent \textbf{\textit{(iii)} \emph{Data Reel} Annotation}: For each selected clip, we recorded the animation types used in the chart, the topic of the clip, the user intent, the number of charts present in the \emph{data reel}, and the corresponding chart types. All annotations were conducted by the authors, who have expertise in data visualization and NLP. To ensure consistent and systematic annotation, we adopted the animation taxonomy proposed by Shi et al., which is specifically designed for real-world data videos \citep{shi2021communicating}. A comprehensive list of the animation types we considered and observed is provided in \Cref{tab:animation_category_distribution}. In addition to animation types, we also annotated the intent of each clip. Here, intent refers to the narrative conveyed by the data reel, as expressed through the combined effect of both the animation and the accompanying narration.

\noindent \textbf{\textit{(iv)} LLM-Based Data Extraction}: Evaluating \emph{data reel} generation requires access to the underlying chart data, which are rarely released with data videos. To address this limitation, annotators captured chart screenshots and used Gemini-2.5-Flash to extract tabular data, following~\citep{masry2025bigcharts}, as demonstrated in (\Cref{fig:chart_data_extract}). We initially extracted data for 360 reels, then automatically flagged and manually reviewed samples with major inconsistencies, such as incorrect chart types or mismatched data-item counts, discarding erroneous cases and retaining 328 reels in the final benchmark. Additionally, we perform a manual audit of 50 tables, 44 were
recovered exactly; the remaining six showed 1--5\% deviations on one or two points, none
reversing a trend, ranking, or relative magnitude. Since the extracted tables serve as inputs for generation and evaluation, preserving major trends and relative magnitudes is more important than exact numerical reconstruction.

Overall, the \datareelbench{} data collection pipeline required substantial human effort. The initial \textbf{Data Video Sourcing} stage required approximately 50 human-hours to review videos across multiple YouTube channels. After channel selection, \textbf{Data Reel Selection} and \textbf{Data Reel Annotation} required approximately 180 human-hours across five annotators. Finally, the \textbf{LLM-Based Data Extraction} stage required an additional 30 human-hours.

\begin{figure*}[t!]
    \centering
    \caption{Overview of the multi-agent framework: (1) The Director Agent converts the input intent, chart data, and duration into a structured animation plan. (2) The Plan Critic Agent reviews the plan to verify that it aligns with the intent. (3) The Coder Agent translates the approved plan into executable HTML and D3.js animation code. (4) The Video Critic Agent evaluates the rendered video to detect issues in timing, narration–animation alignment, or visualization quality, and sends corrections to coder agent for regeneration if needed.}
     \label{fig-MultiAgentic}
    \includegraphics[width=0.98\textwidth]{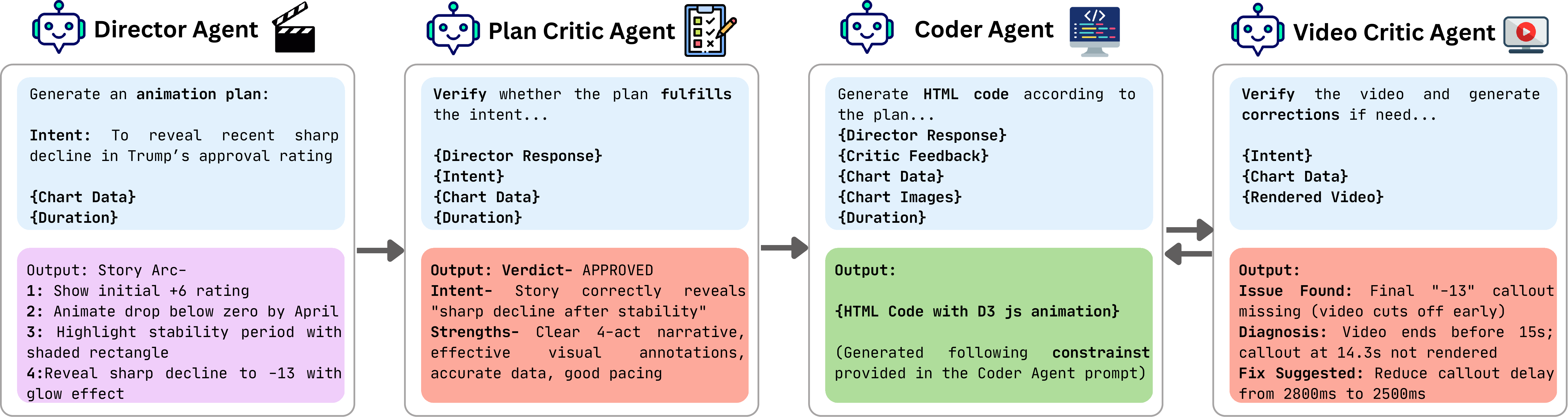} 
    \vspace{-3mm}
\end{figure*}

\vspace{-2mm}

\subsection{Features of \model}

\vspace{-2mm}
We analyze the corpus statistics of \datareelbench{} to highlight its key properties and the challenges it poses for automated data video generation.

\noindent \textbf{Topical and Source Diversity.}
\datareelbench{} spans a wide range of real-world topics, including \emph{Politics \& Policy}, \emph{Business \& Finance}, \emph{Social Issues}, \emph{International Affairs}, and \emph{Environment \& Energy} (\Cref{fig:topic_distribution}), reflecting the broad use of data-driven video storytelling across journalism, education, and public communication. The benchmark aggregates content from 14 distinct YouTube channels (Table~\ref{tab:source_distribution}), providing diverse coverage without reliance on a single dominant source.

\noindent \textbf{Variety of Chart Types and Visual Forms.}
As shown in \Cref{fig:chart_distribution}, \datareelbench{} includes a diverse set of chart types. Bar and line charts are most prevalent, followed by pie charts, area charts, and reels containing multiple chart types.

\noindent \textbf{Rich and Diverse Animation Usage.}
Table~\ref{tab:animation_category_distribution} summarizes the distribution of high-level animation types in the benchmark. Emphasis-driven animations (44.3\%) and suspense-oriented animations (36.4\%) are most common, reflecting the narrative role of animation in guiding viewer attention and pacing information disclosure. Comparison-based animations account for 15.7\%, while more subtle narrative devices such as ellipsis, cohering, and focalization appear less frequently.

\noindent \textbf{Long and Multimodal Narratives.}
\datareelbench{} captures substantial variation in narration length, user intent, and duration. \Cref{fig:tokens_transcript} and \Cref{fig:tokens_intent} show heavy-tailed narration and intent lengths, while \Cref{fig:durations} shows varied durations, with most clips under 20 seconds and a long tail of longer segments.

\vspace{-2mm}
\section{Methodology}
\label{sec:method}

\vspace{-2mm}
\subsection{Task Formulation}
We define the \emph{data reel} generation task as generating, from a given data table, user intent, and target animation duration, and a reference image specifying the visual style, a renderable HTML based on the D3.js library. The final output is the HTML code which consist of a sequence of animated chart animations, paired with narrative subtitles, that together form a coherent data-driven story 
aligned 
with the user’s stated intent. Here, \emph{user intent} refers to the high-level narrative goal that the generated \emph{data reel} should communicate. For example, in the data reel shown in \Cref{fig:example_gpt4o_mini}, the intent is: ``\textit{To demonstrate China's shift from being a major arms buyer to a significant global arms producer and exporter, highlighting its ambition for self-reliance and growing influence}.'' The generated HTML is rendered in a headless browser, where screenshots are captured at regular intervals and assembled into the final video. In real-world \emph{data reels}, such narratives are typically delivered through spoken voiceovers. However, to simplify the task for current models and to avoid challenges related to audio–video synchronization \citep{fernandez2024divas}, we represent narration exclusively through on-screen subtitles in our task.

\vspace{-2mm}
\subsection{\emph{Data Reel} Generation}
We evaluate VLMs under two distinct settings:

\noindent\textbf{\emph{(i)} Single Model Prompting: } In this approach, we benchmark the models using a one-shot generation paradigm. We begin with a minimal initial prompt and iteratively refine it based on the quality of the outputs produced by the models. The final prompt incorporates strict requirements for HTML output formatting, provides guidance on the types of animations to include, and specifies the essential components necessary for correct rendering. Through multiple rounds of prompt refinement, the models progressively generate higher-quality data videos. The prompt for this approach has been shown in \Cref{fig:datavideo_system_prompt_part1} and \Cref{fig:datavideo_system_prompt_part2}.

\noindent\textbf{\emph{(ii)} Multi Agentic Framework:} Recent works have shown LLMs excel at multi-agentic setups \citep{fourney2024magentic, zhu2024teams}, where a number of LLMs are used, each given a certain role to perform a sub-task of an overall complex tasks. Inspired by the applications of such agentic systems in the domain of static story generation \citep{islam-etal-2024-datanarrative}, we propose a multi agentic system for the automated generation of data videos. Existing related systems either focus on static visualizations~\citep{islam-etal-2024-datanarrative, rahman2025text2vis} or take an existing narrative or chart as their starting point rather than a communicative intent~\citep{shen2023dataplayer,ying2024livechart,shao2025narrative}, and so cannot be run on \datareelbench{} without altering the task. We therefore hold the model and inputs fixed and compare generation strategies.

As shown in \Cref{fig-MultiAgentic}, the framework consists of 4 different LLM agents: 

\noindent \textbf{(1) Director Agent}: Given a Data table, video intent, duration and a chart screenshot as input, the director agent is tasked to generate a scene by scene plan with subtitles for the animation.\\
\noindent  \textbf{(2) Plan Critic Agent}: The plan critic agent takes the complete plan and verifies whether the intent is fulfilled, and makes corrections to the plan.\\
\noindent  \textbf{(3) Coder Agent}: Based on the corrected plan, this agent generates HTML code that plays an animation regarding the data.\\
\noindent  \textbf{(4) Video Critic Agent}: This agent evaluates the rendered video to identify and correct visual or temporal issues, such as still frames, misalignment between animations and subtitles, text occlusion, and unintended visual overlaps.

An example produced using this approach is shown in ~\Cref{fig:example_gpt4o_mini}. The prompts for each of the agents are discussed in Appendix \ref{appendix:prompts}.

\vspace{-2mm}

\subsection{Model Selection and Experiment Setup}
\vspace{-1mm}

We evaluate four closed-source and four open-source
 Vision-Language Models (VLMs) on \datareelbench{}, covering diverse dataset sizes and domains. The evaluated models include Gemini 2.5 Pro \citep{comanici2025gemini}, GPT-4.1 Mini \citep{achiam2023gpt}, GPT-5.4 Mini \citep{singh2025openai}, Claude Opus 4.5 \citep{opus4.5}, InternVL 3.5 (8B, 14B) \citep{wang2025internvl3}, and Qwen 2.5 VL (7B, 32B) \citep{bai2025qwen25vltechnicalreport}. All eight models are evaluated under a single-model prompting setup. The best-performing model is then selected for implementing the multi-agent framework.

\begin{table*}[t!]
        \centering
        \caption{HTML-based evaluation of generated code quality across models. Narrative Quality, Informativeness, Subtitle--Transcript Similarity, Code Correctness, and
        Overall are LLM-based scores.}
        \label{tab:htmlEvalNew}

        \setlength{\tabcolsep}{5pt}
        \resizebox{.85\linewidth}{!}{%
        \begin{tabular}{@{}lc|cccccc@{}}
        \toprule
        \textbf{Models}
        & \multicolumn{1}{c|}{\textbf{\makecell{Error\\Rate \%}}}
        & \textbf{\makecell{Narrative\\Quality}}
        & \textbf{\makecell{Informa-\\tiveness}}
        & \textbf{\makecell{Sub.--Trans.\\Similarity}}
        & \textbf{\makecell{Code\\Correctness}}
        & \textbf{Overall}
        & \textbf{\makecell{Std.\\Dev.}} \\
        \midrule

        \rowcolor[HTML]{DBF7FF}
        \multicolumn{2}{@{}l|}{\textit{\faLock \; Closed-Source}} & \multicolumn{6}{l@{}}{} \\
        \rowcolor[HTML]{DBF7FF}
        Gemini~2.5~Pro
        & \multicolumn{1}{c|}{7.67}
        & \textbf{4.66}
        & \textbf{4.63}
        & 3.33
        & \textbf{4.50}
        & \textbf{4.28}
        & 0.87 \\
        \rowcolor[HTML]{DBF7FF}
        GPT~4.1~mini
        & \multicolumn{1}{c|}{6.71}
        & 4.33
        & 4.22
        & \textbf{3.37}
        & 3.52
        & 3.86
        & 0.88 \\
        \rowcolor[HTML]{DBF7FF}
        GPT~5.4~mini
        & \multicolumn{1}{c|}{4.57}
        & 4.49
        & 4.39
        & 2.77
        & 3.52
        & 3.79
        & 0.73 \\
        \rowcolor[HTML]{DBF7FF}
        Claude~Opus~4.5
        & \multicolumn{1}{c|}{\textbf{0.00}}
        & 4.11
        & 4.20
        & 3.24
        & 3.55
        & 3.78
        & 1.25 \\

        \midrule
        \rowcolor[HTML]{E8F5E9}
        \multicolumn{2}{@{}l|}{\textit{\faLockOpen \; Open-Source}} & \multicolumn{6}{l@{}}{} \\
        \rowcolor[HTML]{E8F5E9}
        Qwen~2.5~VL--32B
        & \multicolumn{1}{c|}{12.80}
        & 3.24
        & 3.36
        & 2.43
        & 1.67
        & 2.67
        & 0.82 \\
        \rowcolor[HTML]{E8F5E9}
        InternVL~3.5--14B
        & \multicolumn{1}{c|}{26.30}
        & 1.87
        & 2.34
        & 1.98
        & 0.99
        & 1.79
        & 0.77 \\
        \rowcolor[HTML]{E8F5E9}
        InternVL~3.5--8B
        & \multicolumn{1}{c|}{37.20}
        & 1.26
        & 1.30
        & 1.77
        & 0.80
        & 1.28
        & 0.90 \\
        \rowcolor[HTML]{E8F5E9}
        Qwen~2.5~VL--7B
        & \multicolumn{1}{c|}{39.81}
        & 0.92
        & 0.92
        & 1.37
        & 0.67
        & 0.97
        & 0.79 \\
        \bottomrule
        \end{tabular}
        }

        \vspace{-2mm}
  \end{table*}

\vspace{-2mm}
\subsection{\emph{Data Reel} Evaluation}
\vspace{-1mm}
\label{sec:eval_data_reel}
We evaluate generated \emph{data reels} using two complementary setups. First, we perform LLM-as-a-judge evaluation directly on the generated HTML code to assess code quality across all models. Based on these results, we select the best-performing model, extend it with our multi-agent framework, and render videos from both the single-model and multi-agent outputs. We then conduct pairwise VLM-as-a-judge evaluation on the rendered videos and validate the results through human evaluation. HTML-based evaluation provides an efficient high-level assessment of generated animations, while video-based evaluation offers finer-grained analysis at higher computational cost. For both settings, we use Gemini 3 Pro~\citep{gemini3_1_pro_preview} as the judge.

\noindent\textbf{Code-level evaluation.} Following \citet{islam-etal-2024-datanarrative} and \citet{rahman2025text2vis}, we use an LLM-as-a-judge to score the generated HTML on a 0--5 scale across four criteria inferable from the source: \textit{Narrative Quality} evaluates whether the intended message is conveyed as a coherent, well-structured visual story; \textit{Informativeness} measures the depth of insight in the subtitles and animations, rewarding interpretation beyond restating values; \textit{Subtitle--Transcript Similarity}
assesses how faithfully generated subtitles preserve the meaning of the original narration transcript; and \textit{Code Correctness} examines whether the HTML correctly implements the intended animation. We complement judge-based scoring with \textit{Error Rate}, which measures the percentage of data reels containing JavaScript errors (e.g., page-load and console errors). However, it does not capture animation-related issues such as clipping or subtitle--animation incoherence.

\noindent\textbf{Video centric Evaluation:}
We use VLM-as-a-judge to conduct pairwise comparisons between two videos at a time. This setup enables us to evaluate the multi-agent system against the single-model approach. Our preliminary experiments suggest that VLMs are more reliable in comparative (pairwise) judgments for video data. Based on this insight, we design a blind evaluation protocol in which the VLM is shown two videos without revealing their associated models and is asked to determine which video performs better across three criteria: \textit{Visualization Quality}, \textit{Subtitle--Animation Coherence}, and \textit{Style Consistency}. For each criterion, the judge may select a winner or declare a tie, and the overall winner is determined based on performance across the three categories. Visualization Quality evaluates whether the charts are rendered correctly, remain visually readable, and are free from clipping, overlapping, or rendering artifacts. Subtitle--Animation Coherence assesses whether subtitles are aligned with the visual content, including timing synchronization and semantic consistency. Style Consistency measures how closely each video adheres to the reference visual style in terms of color palette, layout, chart type, and overall design coherence.

\noindent\textbf{Human evaluation:} Human judgment is our primary video-level measure. 
Two co-authors with expertise in data visualization and NLP independently evaluated all 150 pairs using the criteria above, with ties allowed. Annotators were blind to which system produced each video, and the left--right presentation order was randomized per pair.

\vspace{-2mm}
\section{Result Analysis}
\label{sec:results}
\vspace{-2mm}
\subsection{Performance variation in different Models}
\Cref{tab:htmlEvalNew} gives us a look at the performance of the different models. For the 328 samples in the dataset, the metrics represent the average score achieved by different models.  We observe that the closed source models are doing quite well across the four different metrics. Gemini 2.5 pro is the best performer here, followed by GPT 4.1 mini. GPT-5.4 Mini performs similarly to GPT-4.1 Mini overall, except on Subtitle–Transcript Similarity, where GPT-4.1 Mini achieves the best results. Claude Opus, despite being a coding model, falls short on the task. On the other hand, open-source models perform significantly worse. Performance generally improve with increase in model size, but is still significantly behind close source models. One major reason for this poor performance is the size of the HTML files required to properly express the narrative and intent, which often is very near the context window limits of the smaller models. Appendix \ref{app:modelErrors} shows a qualitative analysis of the common errors in different models. This can also been seen from the error rate of the open source models comapred to the close source ones. Small models like Qwen 2.5 VL–7B and InternVL 3.5–8B shows high number of errors at 39.81\% and 37.20\% respectively.

\vspace{-2mm}
\subsection{Agentic vs Direct Generation}
\vspace{-2mm}
 \begin{table*}[t]
  \centering
  \caption{Comparison of human and VLM judgments in the pairwise evaluation of multi-agentic versus single-model outputs. ``Win'' indicates that the multi-agentic output was
  judged to be the better video.}
  \label{tab:blind_eval_results}
  \resizebox{.85\textwidth}{!}{%
  \begin{tabular}{lcccc|cccc|cccc}
  \toprule
  \multirow{2}{*}{\textbf{Component}}
  & \multicolumn{4}{c|}{\textbf{Human$_1$}}
  & \multicolumn{4}{c|}{\textbf{Human$_2$}}
  & \multicolumn{4}{c}{\textbf{VLM}} \\

  \cmidrule(lr){2-5} \cmidrule(lr){6-9} \cmidrule(lr){10-13}
   & \textbf{Win} & \textbf{Loss} & \textbf{Tie} & \textbf{p-value}
   & \textbf{Win} & \textbf{Loss} & \textbf{Tie} & \textbf{p-value}
   & \textbf{Win} & \textbf{Loss} & \textbf{Tie} & \textbf{p-value} \\

  \midrule
  Visualization Quality
  & \textbf{84} & 42 & 24 & $\bm{2.30e^{-4}}$
  & \textbf{75} & 49 & 26 & $\bm{2.44e^{-2}}$
  & \textbf{103} & 21 & 26 & $\bm{3.50e^{-14}}$ \\

  Sub. Ani. Coherence
  & \textbf{80} & 44 & 26 & $\bm{1.56e^{-3}}$
  & \textbf{61} & 43 & 46 & $9.50e^{-2}$
  & \textbf{70} & 14 & 66 & $\bm{4.07e^{-10}}$ \\

  Style Consistency
  & \textbf{132} & 13 & 5 & $\bm{5.74e^{-26}}$
  & \textbf{136} & 8 & 6 & $\bm{3.58e^{-31}}$
  & \textbf{148} & 2 & 0 & $\bm{1.59e^{-41}}$ \\

  \midrule
  \textbf{Overall}
  & \textbf{111} & 32 & 7 & $\bm{2.07e^{-11}}$
  & \textbf{108} & 36 & 6 & $\bm{1.50e^{-9}}$
  & \textbf{138} & 12 & 0 & $\bm{2.64e^{-28}}$ \\

  \bottomrule
  \end{tabular}
  }
  \vspace{-2mm}
  \end{table*}
We next ask whether structured generation can alleviate these challenges. Using Gemini~2.5 Pro as the underlying model, we compare direct prompting with our multi-agent framework on 150 randomly selected data reels. Each pair of rendered videos is evaluated independently by two human judges and a VLM judge using blind pairwise comparison.

As shown in Table~\ref{tab:blind_eval_results}, both human judges consistently prefer the multi-agent framework, selecting it as the better overall video in 111/150 and 108/150 comparisons, respectively. The VLM judge exhibits an even stronger preference (138/150). Gains are largest in \textit{Style
Consistency} and are statistically significant for both human judges in \textit{Visualization Quality}. \textit{Subtitle–Animation Coherence} improves less consistently across judges, indicating that temporal synchronization remains a challenging aspect of the task. The two human judges agree on 81.3\% agreement (Cohen's $\kappa=0.58$), while the VLM judge agrees with each human in 76.7\% and 74.7\% of cases.
The large gains in \textit{Style Consistency} are primarily driven by the planning and verification stages of the multi-agent framework. Compared with direct prompting, the generated videos more faithfully reproduce the reference layouts, color palettes, and design elements, whereas single-prompt generation often simplifies the visual design and overlooks stylistic details.

\vspace{-2mm}

\subsection{Ablation Study}
\label{sec:ablation}

\vspace{-2mm}
To further assess the contribution of each component, we conduct ablation studies by removing the critic agents from the multi-agent pipeline. This decreases performance compared to the full framework, suggesting that the critic agents play important roles in improving the quality of generated data reels. Detailed results are provided in Appendix~\ref{app:noCritic}.

\vspace{-2mm}
\subsection{Qualitative Analysis}
\label{main:qua}

To better understand the remaining challenges in generated videos, we randomly sampled 25 successfully renderable data reels from each model and manually inspected their outputs. We identify four major recurring errors across the models:

\noindent \textbf{Reels with no Animation}: A common failure case, especially for open-source models and smaller closed-source models such as GPT-4.1 Mini and GPT-5.4 Mini, is that the generated HTML contains only a static chart without visible animation or subtitles. Examples of this issue are shown in \Cref{fig:gpt}(a), \Cref{fig:gpt56}, and \Cref{fig:open}.

\noindent \textbf{Lack of synchronization with subtitles}: Another frequent issue is poor coherence between animations and subtitles. In several cases, the animation starts much later than the corresponding subtitle, as shown in \Cref{fig:gpt}(c) and \Cref{fig:gpt}(d). We also observe pacing issues, where models struggle to align animation speed and subtitle timing. 

\noindent \textbf{Overlapping and clipped elements}: Generated reels often contain overlapping or clipped visual elements, including text, axes, and legends. As illustrated in \Cref{fig:claude}(a), these issues make the resulting \emph{data reels} difficult to interpret.

\noindent \textbf{Lack of adherence to the reference}: The single-prompt approach often fails to follow the provided reference image. This is also reflected in \Cref{tab:blind_eval_results}, where both human and VLM judges prefer the multi-agentic approach for style consistency. Similar failures are observed for other models as well; for example, outputs from Claude Opus, shown in \Cref{fig:claude}, frequently use a purple background regardless of the provided reference.

\noindent \textbf{Additional rendering failures}: 
We discuss additional errors, including external component failures, and improper chart positioning in Appendix~\ref{app:qua}.

\vspace{-2mm}

\section{Conclusion}
\vspace{-2mm}
In this paper, we introduced \datareelbench{}, a benchmark for automated generation of animated data-driven video stories that require coordinating data, visualization, animation, and narration. We also include a baseline based on a multi-agent framework that decomposes \emph{data reel} generation into planning, generation, and verification stages. Our automatic and human evaluations show that this baseline improves over single-model prompting, particularly in style consistency and overall visual quality, while also revealing remaining challenges. Future extensions of this work can use the  benchmark to develop better models and more reliable evaluation methods. In particular, supervised fine-tuning of smaller open-source models on curated examples of \emph{data reel} plans, animation code, and subtitle-aligned visual transitions may improve their ability to generate coherent data videos. We hope \datareelbench{} provides a useful foundation for future research on automated data video generation.

\section*{Limitations}

\noindent\textbf{Scope.} Our benchmark targets \emph{data reels}: the chart-centric
animated segments of data videos, generated as executable code with on-screen subtitles. We
isolate chart-centric animation deliberately, since it constitutes the core
visualization-generation problem while holding constant the distinct challenges of speech
synthesis and cinematic editing; photographic B-roll, iconography, and narration audio
therefore fall outside the task definition (\S\ref{sec:method}) rather than being omissions
from it. Representing narration as subtitles does simplify a genuine dimension of the
problem, since audio--animation synchronization and pacing are central to professional data
videos \citep{fernandez2024divas}. All transcripts and intents are in English. Extending
\datareelbench{} to spoken and multilingual narration is clear future work.

\noindent\textbf{Benchmark scale.} Like Design2Code (484 examples;
\citealp{si-etal-2025-design2code}), Plot2Code (132; \citealp{wu-etal-2025-plot2code}), and
many recent code-generation benchmarks at ACL venues, \datareelbench{} (328 reels) is
intended primarily as an evaluation benchmark rather than a large-scale training corpus,
and is sized for evaluation reliability, affordability, and community adoption. Two costs
bound it. Annotation is a search problem: reels sit buried inside long-form videos, and
each retained reel required roughly 15 minutes of review and 15 minutes of
expert annotation covering timestamping, transcription, animation and intent labeling, and
table verification. The full benchmark-construction pipeline required approximately 260 human-hours overall, including 180 hours for reel selection and annotation. Evaluation is costlier
still: where adding a sample to a text benchmark costs an API call, every \datareelbench{}
sample must be generated, rendered, and watched, at roughly \$ 0.20 per sample and
\$ 1.5 under the multi-agent baseline, whose Video Critic consumes rendered frames.
Scaling further, particularly to support supervised fine-tuning, remains future work.

\noindent\textbf{Diversity.}
\datareelbench{} prioritizes diversity along the dimensions the task turns on: animation
behavior, chart type, topic, and intent. Coverage across source channels is less uniform,
since channels differ by an order of magnitude in how many qualifying reels they publish,
and enforcing uniform counts would have meant discarding strong reels or padding with weak
ones.

\noindent\textbf{Data extraction.}
Underlying tables are rarely published with data videos, so we recover them from chart
screenshots using Gemini-2.5-Flash. Because the extracted table is both the model input and the evaluation reference, extraction errors do not affect the fairness or comparability of benchmark evaluation. Absolute values may
nonetheless deviate from the original broadcasts, so \datareelbench{} should not be treated as a source of factual statistics.

\noindent\textbf{Evaluation methodology.}
Evaluating \emph{data reels} requires judging visualization quality,
subtitle--animation coherence, and style consistency jointly, and these criteria retain a
preference-dependent component even under a rubric.
Automatic evaluation does not yet substitute for
human judgement (\S\ref{sec:eval_data_reel}), so benchmarking on \datareelbench{} currently carries a
human-annotation cost, though one paid per model evaluated rather than per training
iteration. These constraints primarily bound how broadly our findings generalize rather
than the validity of the comparisons we report.

\section*{Ethics Statement}
This study explores the capabilities of VLMs for automated data video clip, or \emph{data reel}, generation. The curated benchmark is collected from publicly available YouTube channels. The benchmark does not include the full data video clips; instead, it provides only the video links and timestamps corresponding to the relevant data video segments. Since the benchmark is based on publicly accessible online content, no private or sensitive user data is collected. Data annotation was conducted by the authors of this study; therefore, no external annotators were involved or compensated. Finally, we used AI-based writing assistants only to improve the presentation of paper.

\bibliographystyle{acl_natbib}
\bibliography{ms}

\input{app}

\end{document}

%% file: math-com.tex
\usepackage{amsmath}
\usepackage{amsfonts,bm}
\usepackage{xspace}

\definecolor{mypink3}{cmyk}{0, 0.7808, 0.4429, 0.1412}

\makeatletter   
\newcommand{\sveryshortarrow}[1][3pt]{\mathrel{%
    \vcenter{\hbox{\rule[-.5\fontdimen8\scriptfont3]
               {\scriptratio\dimexpr#1\relax}{\fontdimen8\scriptfont3}}}%
   \mkern-4mu\hbox{\let\f@size\sf@size\usefont{U}{lasy}{m}{n}\symbol{41}}}}
\makeatother

\def\eqref#1{equation~\ref{#1}}

\def\1{\bm{1}}

\def\m1{{\bm{1}}}

\DeclareMathAlphabet{\mathsfit}{\encodingdefault}{\sfdefault}{m}{sl}
\SetMathAlphabet{\mathsfit}{bold}{\encodingdefault}{\sfdefault}{bx}{n}



%% file: app.tex
\renewcommand{\datareelbench}{
\textsc{
\textcolor{datareelbench1}{Data}\textcolor{datareelbench2}{Reel}
}
}
\appendix
\begin{appendices}

\begin{figure*}[h]
     \centering
        \includegraphics[width=\textwidth]{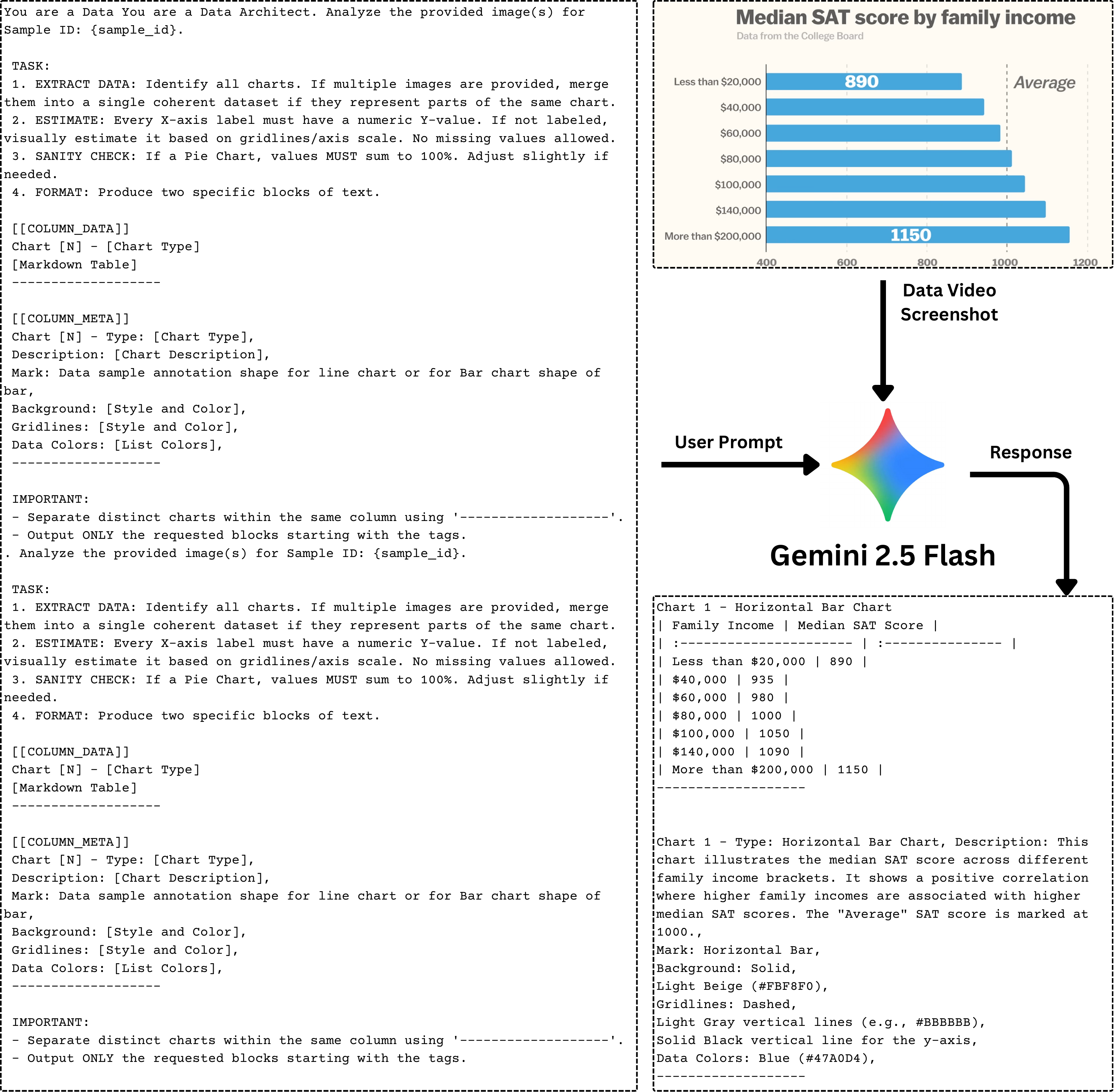}
         \caption{
         The figure presents an overview of the chart data extraction process using the Gemini-2.5-flash \cite{geminiteam2023gemini} model. 
          }
    \label{fig:chart_data_extract}
\end{figure*}

\section{Chart Data Extraction}
\label{app:chart_data_ext}

We utilize the multi-modal large language model (MLLM) Gemini-2.5-Flash \cite{geminiteam2023gemini} to extract data from chart images. The model was given a screenshot of the chart that we collected during our data collection stage. The annotators were instructed to collect the screenshot when the entire chart was visible. In case the animations occluded particular portions of the chart or if the same clip consisted of multiple animated segments, multiple images were collected and they all were sent to the model together. The models identified both the data present in the chart and associated meta-data like colors of different portions, annotation, etc. The detailed overview of the process is shown in \Cref{fig:chart_data_extract}

\section{Data Sources}

Table \ref{tab:source_distribution} shows the distribution of \datareelbench by Source Channels. Every channel has their own animation style, which makes the tasks of \emph{data reel} generation a difficult task as it also involves the replication of the animation type used by the channel.

\begin{table}[t]
\centering
\small
\setlength{\tabcolsep}{6pt}
\renewcommand{\arraystretch}{1.1}
\caption{\textbf{Distribution of Data Reels by Source Channel}. Counts are followed by percentages in parentheses.}
\label{tab:channel_distribution}
\begin{tabular}{l r}
\toprule
\textbf{Channel} & \textbf{\# Reels (\%)} \\
\midrule
Vox & 112 (34.1\%) \\
The Wall Street Journal & 82 (25.0\%) \\
TLDR News Global & 41 (12.5\%) \\
Economics Explained & 30 (9.1\%) \\
Polymatter & 23 (7.0\%) \\
TLDR News & 19 (5.8\%) \\
The Guardian & 6 (1.8\%) \\
Kurzgesagt -- In a Nutshell & 4 (1.2\%) \\
The Gravel Institute & 3 (0.9\%) \\
Our Changing Climate & 2 (0.6\%) \\
CBC News & 2 (0.6\%) \\
The Infographics Show & 2 (0.6\%) \\
Politizane & 1 (0.3\%) \\
TLDR News EU & 1 (0.3\%) \\
\bottomrule
\end{tabular}
\label{tab:source_distribution}
\end{table}

\begin{figure*}[t]
    \centering
    \begin{subfigure}[t]{0.48\columnwidth}
        \centering
        \includegraphics[width=\linewidth]{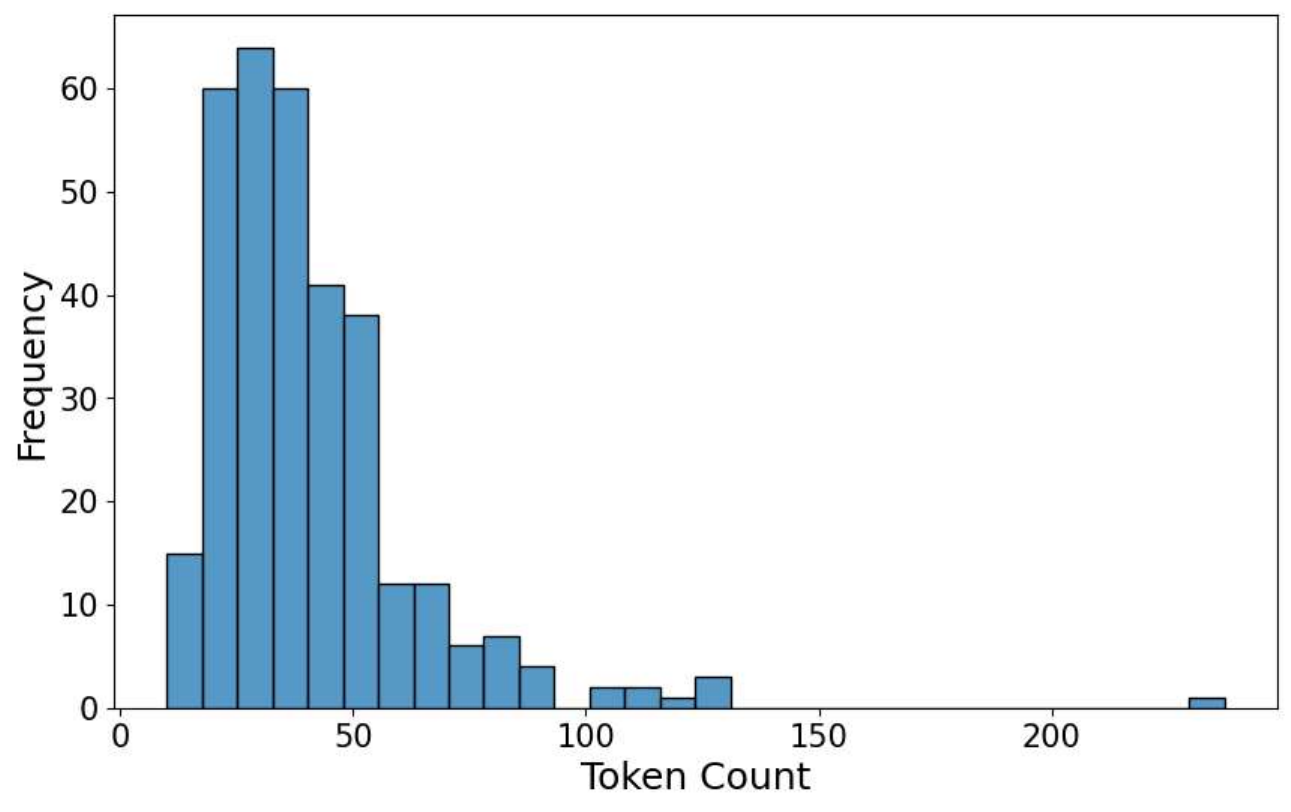}
        \caption{Transcript length (tokens).}
        \label{fig:tokens_transcript}
    \end{subfigure}
    \hfill
    \begin{subfigure}[t]{0.48\columnwidth}
        \centering
        \includegraphics[width=\linewidth]{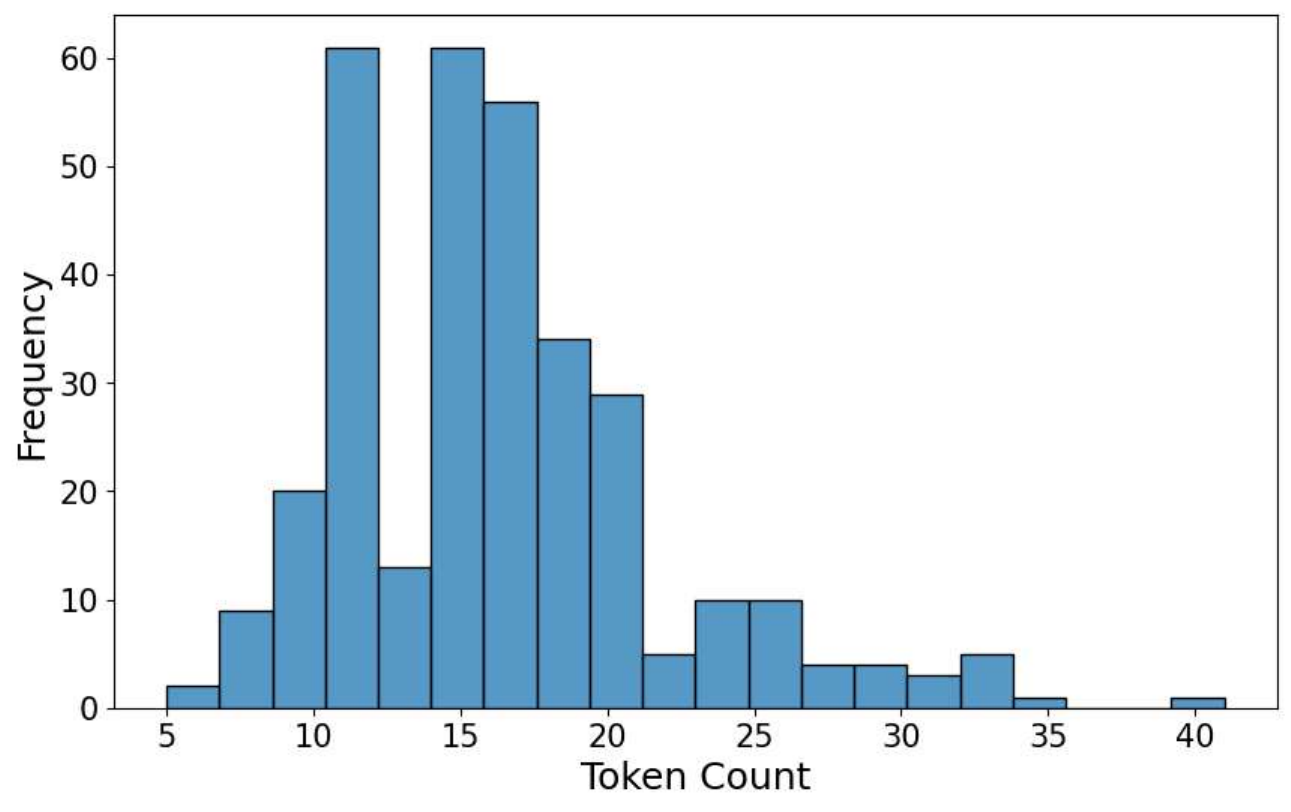}
        \caption{Intent description length (tokens).}
        \label{fig:tokens_intent}
    \end{subfigure}
    \caption{\textbf{Token distribution statistics.} Distribution of token counts for (a) narration transcripts and (b) intent descriptions.}
    \label{fig:token_distribution}
\end{figure*}

\section{Additional Benchmark Details}
\label{app:AddlDetails}
\Cref{fig:token_distribution}a and \Cref{fig:tokens_intent} show the token count distribution of the transcript and intent of the collected \emph{data reels}.

\begin{table}[t]
\centering
\small
\setlength{\tabcolsep}{6pt}
\renewcommand{\arraystretch}{1.1}
\caption{\textbf{Distribution of Main Animation Categories}. Counts are followed by percentages in parentheses.}
\label{tab:animation_category_distribution}
\begin{tabular}{l r}
\toprule
\textbf{Animation Category} & \textbf{\# Reels (\%)} \\
\midrule
Emphasis & 297 (44.3\%) \\
Suspense & 244 (36.4\%) \\
Comparison & 105 (15.7\%) \\
Ellipsis & 17 (2.5\%) \\
Cohering & 4 (0.6\%) \\
Focalization & 3 (0.4\%) \\
\bottomrule
\end{tabular}
\end{table}

\section{Prompts}
\label{appendix:prompts}

Prompt engineering was an important part of the automated \emph{data reel} generation process. Every prompt was tuned for a number of times to fit the required task being performed. The same prompt was used for different models when performing the same task. 

For the single model approach of data reel generation, the prompt used has been shown in \Cref{fig:datavideo_system_prompt_part1} and \Cref{fig:datavideo_system_prompt_part2}. Apart from the single-model approach, we also employ a multi-agent framework for \emph{data reel} generation. The prompts used by the different agents in this framework are shown in \Cref{fig:director_prompt} (Director Agent), \Cref{fig:plan_critic_prompt} (Plan Critic Agent), \Cref{fig:coder_prompt} (Coder Agent), and \Cref{fig:video_critic_prompt} (Video Critic Agent). The prompts used for the HTML-based LLM judge and the pairwise video VLM judge are shown in \Cref{fig:html_judge_prompt} and \Cref{fig:pairwise_vlm_judge_prompt}, respectively. For some figures, the end portion of the prompt detailing the output structure had to be truncated to fit on one page.

\section{Ablation Study}
\label{app:noCritic}
\begin{table}[t]
\resizebox{\linewidth}{!}{%
\centering
\caption{Ablation study comparing the agentic pipeline without critic components (No-Critic) against the Single-Shot baseline across evaluation components.}
\label{tab:ablation_no_critic}

\begin{tabular}{lccc}
\toprule
\textbf{Component} 
& \textbf{Wins} 
& \textbf{Loss} 
& \textbf{Ties} \\
\midrule
Visualization Quality        & 67  & 51 & 32 \\
Subtitle--Animation Coherence & 50  & 44 & 56 \\
Style Consistency            & 147 & 0  & 3 \\
\midrule
\textbf{Overall}             & 117 & 33 & 0 \\
\bottomrule
\end{tabular}
}
\end{table}

In order to verify the importance of the two critic agents, we conducted an ablation study where we used only the Director agent and Coder agent for the purpose of \emph{data reel} generation. We then compared the results with the ones generated by the single agent approach using the VLM judge. The results have been shown in \Cref{tab:ablation_no_critic}. We observe that, compared to \Cref{tab:blind_eval_results}, the judge prefers the two-agent approach 117 times, which was previously 138 when all four agents were used. The addition of the critic agents improve the Visualization quality of the agentic workflow, as it increases from 67 wins with only director and coder agent to 103 wins out of 150 when the critic agents are used. Style consistency remains same for both cases, whereas Subtitle–Animation Coherence also sees a slight improvement.

\begin{figure*}[t]
    \centering
    \caption{Overview of our benchmark construction pipeline: (1) We first identify videos that contain \emph{data reels} from YouTube channels that regularly publish data videos. (2) We identify the timestamp of the \emph{data reel} from each of the selected videos along with the transcript. (3) Annotators then annotate animation related metadata such as animation type and user intent. (4) LLMs are used to extract chart data from the \emph{data reels} in tabular format.}
     \label{fig:Benchmark}
    \includegraphics[width=0.98\textwidth]{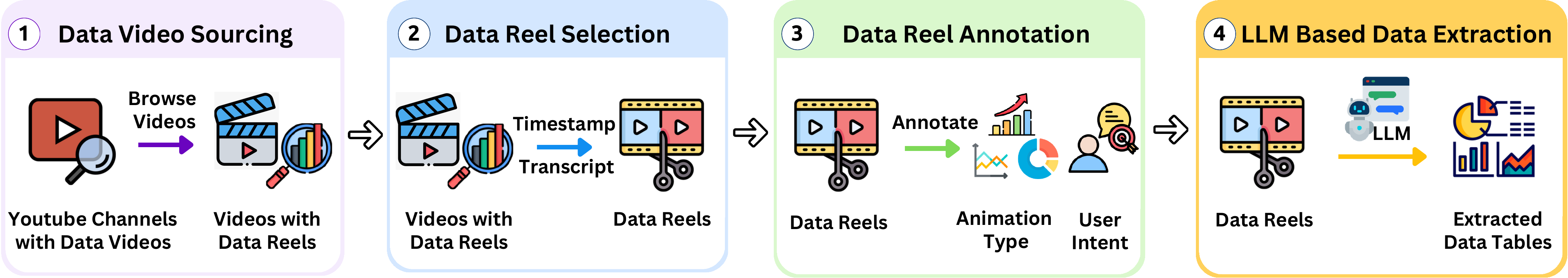} 
    \vspace{-2mm}
\end{figure*}

\section{Qualitative Analysis}
\label{app:qua}

As discussed previously, from each of the models we examined, we randomly sampled 25 \emph{data reels} and identified existing issues within them. Some additional issues have been discussed below in detail:

\noindent \textbf{External Component errors}: When asked to follow the reference, it was observed that some models \Cref{fig:gpt}(b), specially the multi-agentic approach tried utilizing external objects like pictures as background, icons etc. While this worked for most files, some cases were observed where the models hallucinated image URL links, which ultimately did not render the desired background for the reels.

\noindent \textbf{Instability in Animations and Charts}: In some cases, the animations or charts appear jittery. For example, in \Cref{fig:claude}(c) \& \Cref{fig:claude}(d) the visual elements appear to turn on and off intermittently, which disrupts the continuity of the display. Whenever the system attempts to remove focus from a specific part of the chart, the entire visualization tends to jitter and behave as if the whole graph is restarting, rather than smoothly unfocusing only the intended component.

\noindent \textbf{Improper Chart Positioning}: In several instances like \Cref{fig:gemini}, the generated charts appear at distant or unexpected locations on the screen rather than near the center of the frame. This misplacement makes the visualization difficult to focus on and reduces its overall clarity. Similarly, subtitles also appear in unintended positions, misaligned with the charts, further reducing the readability and effectiveness of the presentation.

\section{Examples of a Reference Image}

In this section, we demonstrate a reference image which is used as an input for the datareel generated in \Cref{fig:example_gpt4o_mini}. The reference image is a screenshot of the chart shown in the original data video.

\begin{figure}[t]
     \centering
     \includegraphics[width=.97\columnwidth]{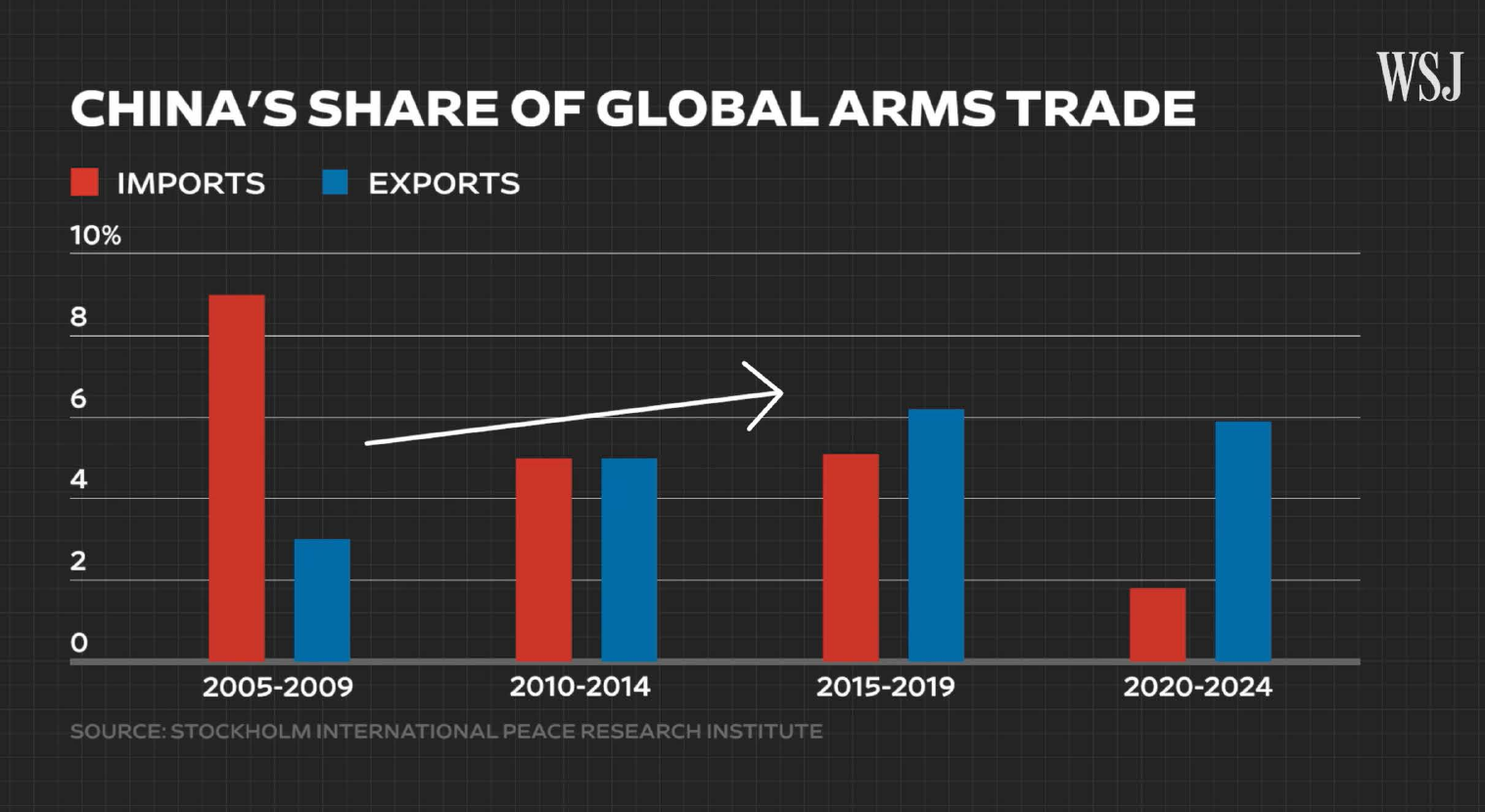}
     \caption{
        Reference image for DataReel generated in \Cref{fig:example_gpt4o_mini}.
     }
     \label{fig:RefExample}
\vspace{-3mm}     
\end{figure}

\section{Error Cases from Different Models}

\label{app:modelErrors}
\begin{figure*}[t]
    \centering
    \includegraphics[scale=0.39]{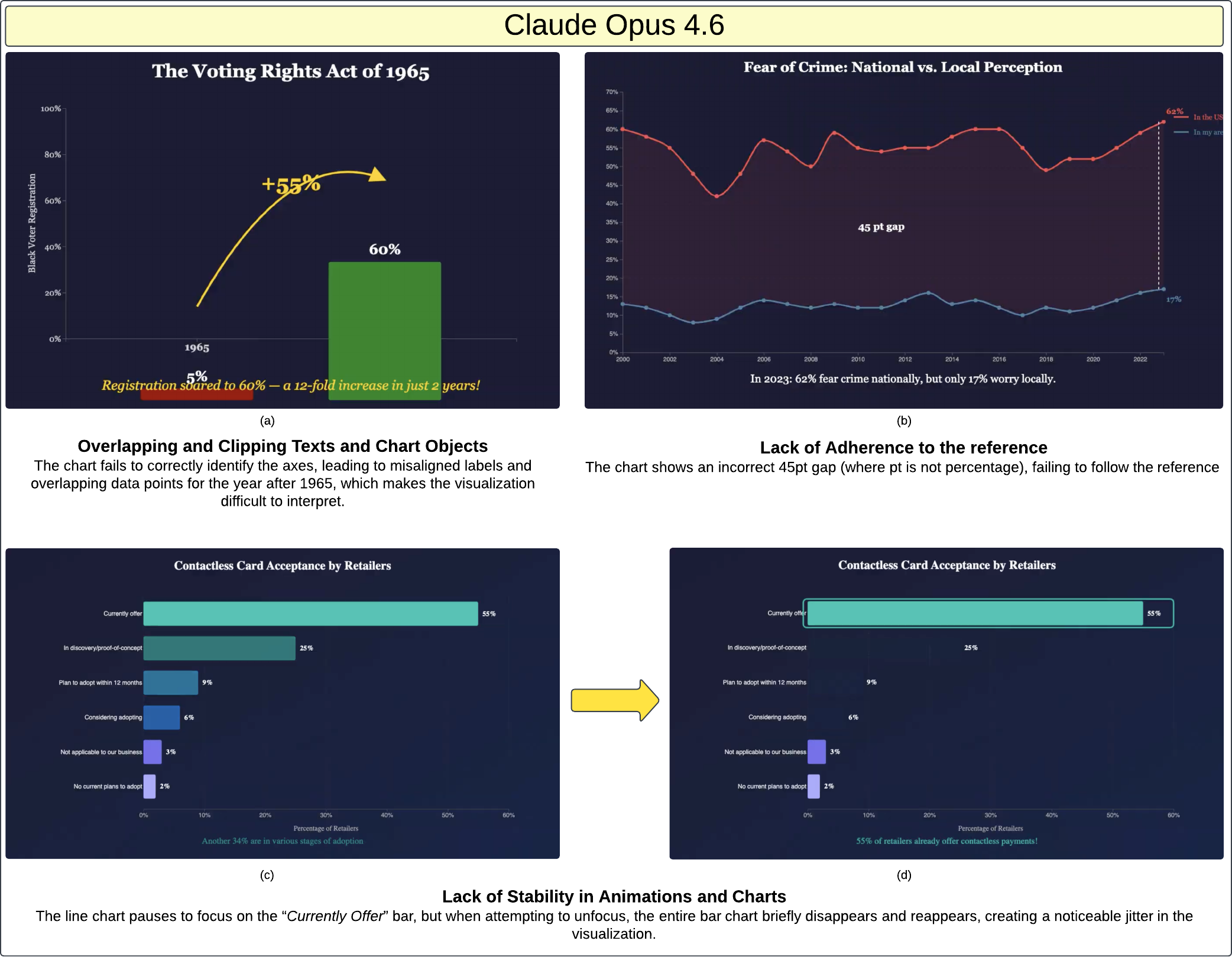}
    \caption{Claude Opus 4.5 exhibits overlapping chart elements, incorrect axis interpretation, an erroneous 45pt gap that does not correspond to the reference, and unstable animations where the chart briefly disappears and reappears when shifting focus}
    \label{fig:claude}
\end{figure*}

\begin{figure*}[t]
    \centering
    \includegraphics[scale=0.43]{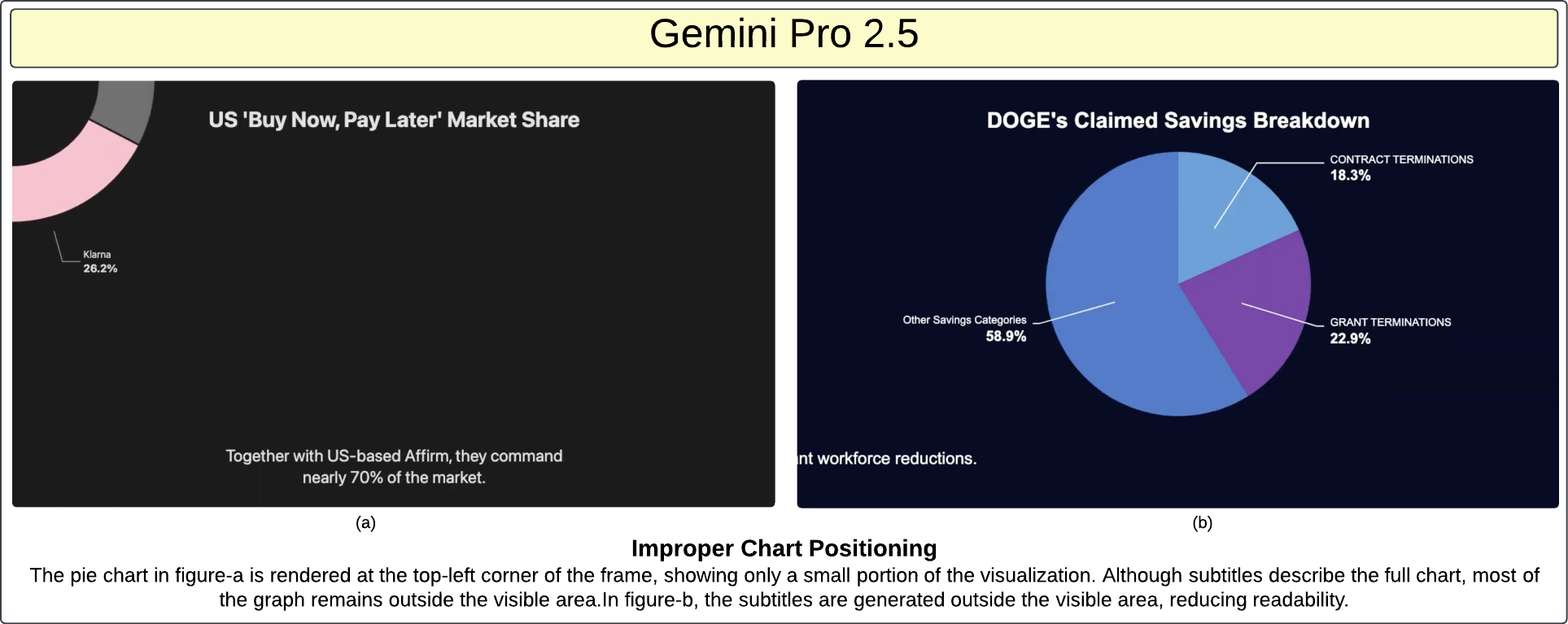}
    \caption{Gemini Pro 2.5 demonstrates improper chart positioning where the pie chart is partially rendered at the top-left corner and subtitles are sometimes generated outside the visible frame. }
    \label{fig:gemini}
\end{figure*}

\begin{figure*}[t]
    \centering
    \includegraphics[scale=0.43]{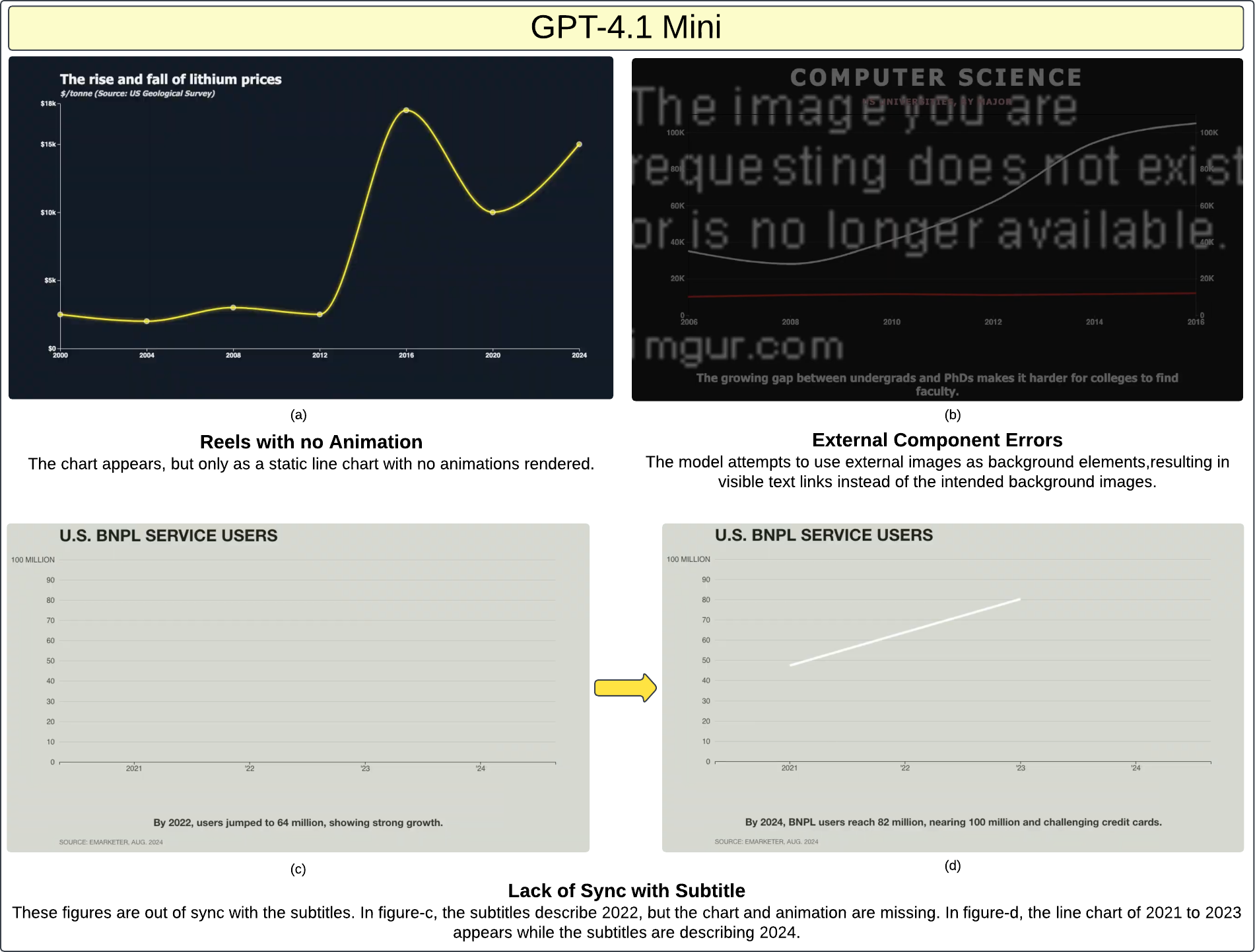}
  \caption{GPT-4.1 mini shows missing animations and synchronization issues, where only static charts are generated and the chart updates appear out of sync with the subtitles. External component errors are also observed, such as failed loading of background images due to invalid or hallucinated URLs.}
    \label{fig:gpt}
\end{figure*}

\begin{figure*}[t]
    \centering
    \includegraphics[scale=0.41]{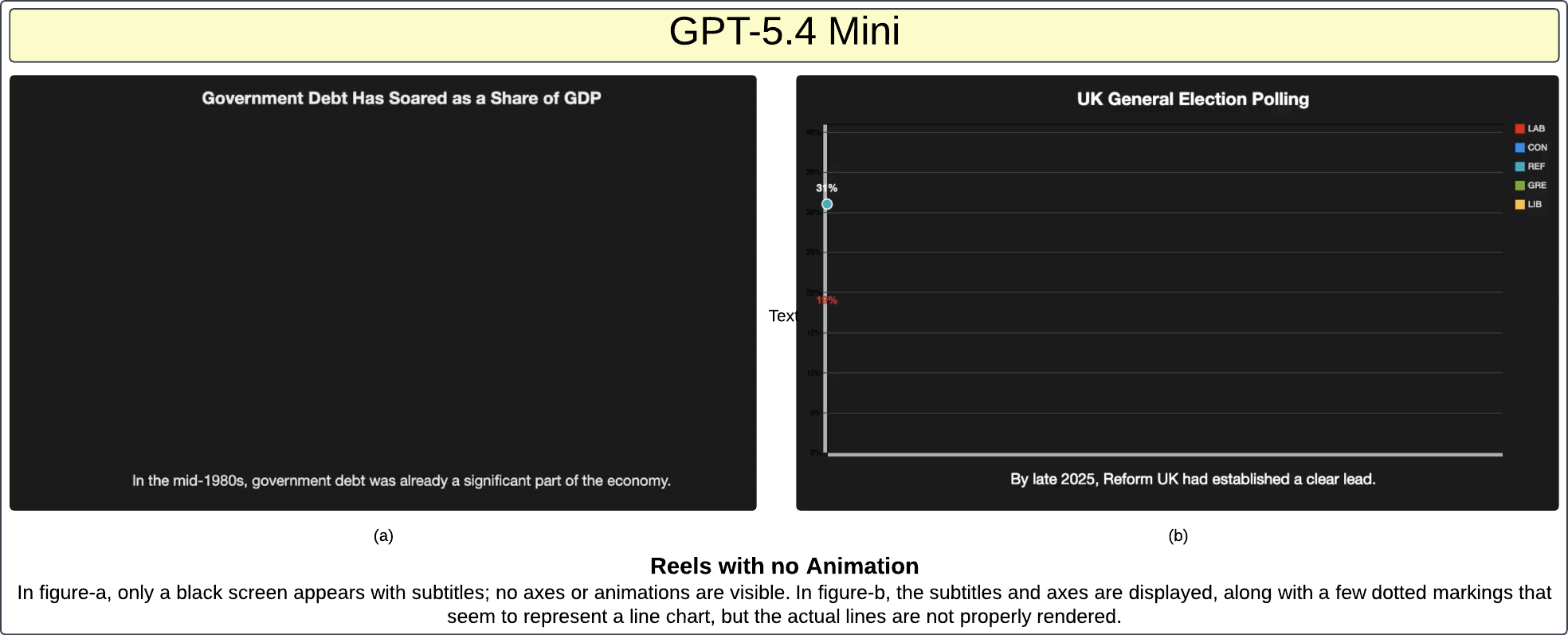}
    \caption{Examples of failure cases from ChatGPT 5.4 Mini during chart visualization tasks. In figure-(a), the result shows only a black screen accompanied by subtitles, with no visible axes or graphical elements. In figure-(b), the axes and subtitles appear, but the chart is improperly rendered with scattered dotted points instead of coherent line structures, resulting in an incomplete and ambiguous representation.}
    \label{fig:gpt56}
\end{figure*}

\begin{figure*}[t]
    \centering
    \includegraphics[scale=0.42]{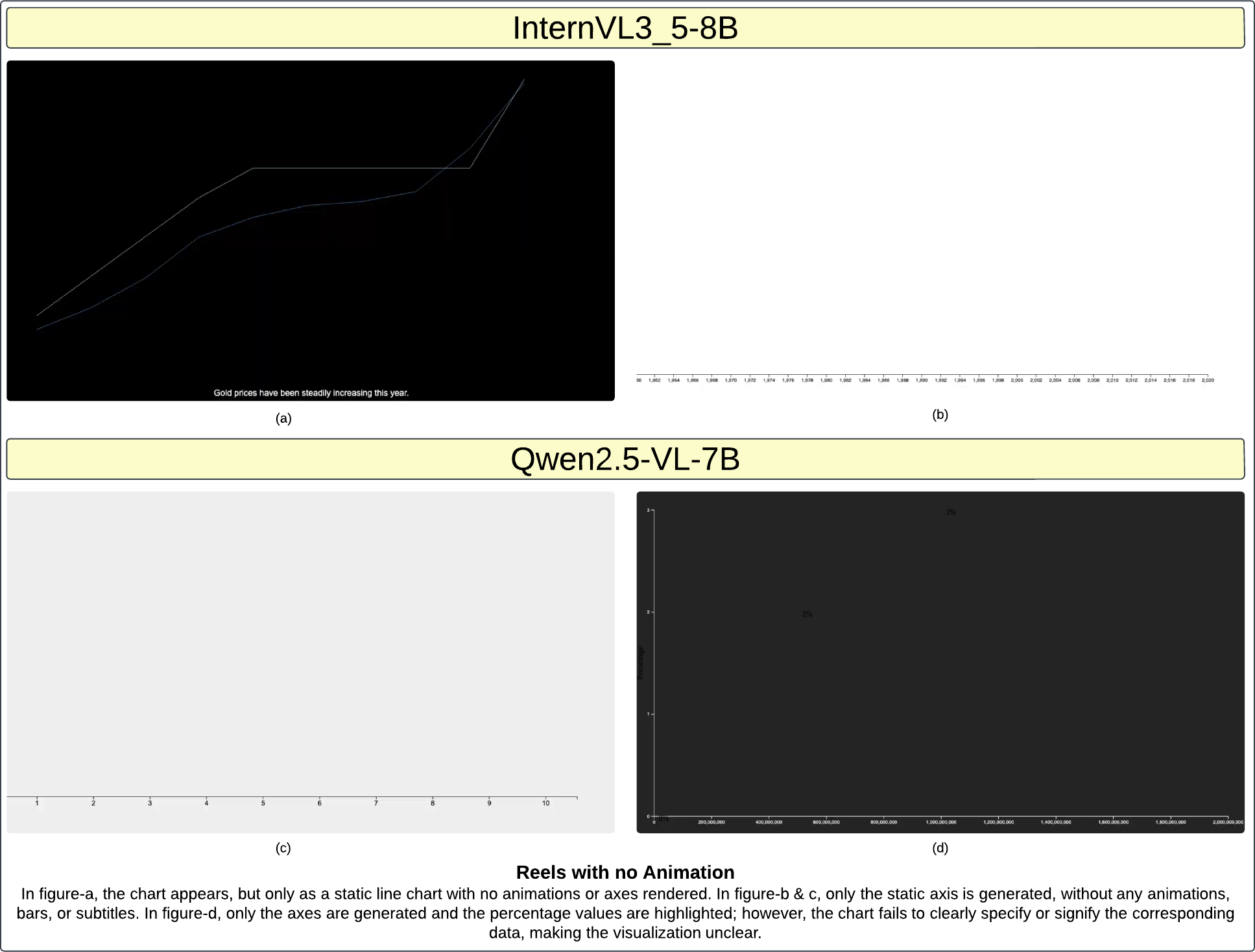}
    \caption{Failure cases from open-source vision–language models when generating chart-based reel content. (a–b) Outputs from InternVL3.5-8B and (c–d) outputs from Qwen2.5-VL-7B. Instead of producing a complete animated chart, the models generate incomplete visualizations such as static axes, partial lines, or highlighted percentages without proper data representation, resulting in unclear or blank diagrams.}
    \label{fig:open}
\end{figure*}

We present common errors that were found while manually going through the outputs from different models during one shot \emph{data reel} generation. The error cases from the models Claude-Opus-4.5, Gemini-2.5-pro, GPT-4.1-mini, GPT-5.4-mini and the 2 open source models (InternVL3.5-8B and Qwen2.5-VL-7B) have been shown in \Cref{fig:claude}, \Cref{fig:gemini}, \Cref{fig:gpt}, \Cref{fig:gpt56} and \Cref{fig:open} respectively.

\begin{figure*}[t]
    \centering
    
    \begin{tcolorbox}[
        colback=gray!5!white,
        colframe=gray!60!black,
        title=\textbf{Data Video Generator System Prompt (Part 1)},
        fonttitle=\bfseries\small,
        boxrule=0.8pt,
        arc=2pt,
        left=2pt, right=2pt, top=2pt, bottom=2pt
    ]
    \ttfamily\scriptsize
    \sloppy
    \setlength{\emergencystretch}{3em}

    You are generating a **self-contained HTML file** that animates one or more charts \\
    for a short **data video scene** using D3 js. \\

    The goal is to create an **animated visual story** that fulfills the stated \\
    **intent of the scene**, using ONLY the provided data. \\

    The input may contain MULTIPLE charts. \\
    All charts belong to the SAME scene and must be animated within ONE SVG. \\

    ================================================== \\
    CORE REQUIREMENT (VERY IMPORTANT) \\
    ================================================== \\

    - Author a cohesive STORY that conveys the intent using animation. \\
    - The story must be constructed from the provided data ONLY. \\
    - The narrative must be expressed through **on-screen subtitles** and **visual animation**. \\
    - The animation + subtitles together must clearly fulfill the stated **intent**. \\
    - Subtitles should explain, highlight, compare, or summarize the data as needed. \\
    - The story should unfold progressively over time, not all at once. \\

    ================================================== \\
    STRICT RULES \\
    ================================================== \\

    - Output ONLY valid HTML \\
    - Use exactly ONE <svg id="chart"> \\
    - The SVG represents the FULL video frame \\
    - SVG MUST use a fixed resolution of 1280x720 pixels \\
    - <svg id="chart" width="1280" height="720"> \\
    - Do NOT resize the SVG dynamically \\

    - Define: \texttt{window.\_\_VIDEO\_DURATION\_\_} = \{duration\} \\
    - THIS TIMING IS STRICTLY ENFORCED DURING RENDERING \\
    - Use the entire duration meaningfully \\
    - Allow the story to unfold progressively \\
    - Avoid finishing too early or rushing key moments \\

    - Define functions: resetChart() and scheduleNow() \\
    - scheduleNow() MUST produce a complete animation every time it is called \\
    - resetChart() MUST restore the SVG to a valid initial state \\
    - Use setTimeout for ALL animations \\
    - NEVER reference requestAnimationFrame (directly or indirectly) \\
    - Do NOT load external assets except the D3 CDN \\
    - Do NOT invent, interpolate, or modify data values \\
    - Deterministic animation only (no randomness, no frame-based callbacks) \\

    ================================================== \\
    SUBTITLE REQUIREMENTS (MANDATORY) \\
    ================================================== \\

    - You MUST generate subtitles to narrate the story \\
    - Substitles for a scene must be long enough to be readable by the viewer \\
    - Subtitles must appear within the SVG (not HTML overlays) \\
    - Subtitles must be synchronized with animation events \\
    - Subtitles must explain the story implied by the intent \\
    - Subtitles must be readable, non-overlapping, and stay within SVG bounds \\
    - Subtitles MUST fit entirely inside the 1280x720 frame at all times \\
    - Subtitle text must wrap or line-break to avoid overflow \\
    - Subtitles must never be clipped or cropped \\
    - Use a consistent subtitle position (e.g., bottom-center) \\

    \end{tcolorbox}

    \caption{Data Video Generator system prompt — core requirements and strict execution rules.}
    \label{fig:datavideo_system_prompt_part1}
\end{figure*}

\begin{figure*}[t]
    \centering
    
    \begin{tcolorbox}[
        colback=gray!5!white,
        colframe=gray!60!black,
        title=\textbf{Data Video Generator System Prompt (Part 2)},
        fonttitle=\bfseries\small,
        boxrule=0.8pt,
        arc=2pt,
        left=2pt, right=2pt, top=2pt, bottom=2pt
    ]
    \ttfamily\scriptsize
    \sloppy
    \setlength{\emergencystretch}{3em}

    ================================================== \\
    LAYOUT \& LEGIBILITY CONSTRAINTS \\
    ================================================== \\

    - All charts, subtitles, axes, labels, and annotations MUST fit within 1280x720 \\
    - Use explicit inner margins; do NOT place marks on SVG edges \\
    - Reserve space for subtitles and annotations \\
    - Ensure no overlap between visual elements \\
    - Axis labels, tick labels, annotations, and subtitles must not collide \\
    - Prioritize clarity and readability over visual flair \\

    ================================================== \\
    NARRATIVE ANIMATION STRATEGIES (GUIDANCE) \\
    ================================================== \\

    Choose animation strategies that best support the **intent** and the story you \\
    are constructing from the data: \\

    - **Emphasis** (highlight key values, peaks, outliers, or categories) \\
    - **Suspense** (gradual reveal, delayed comparison, count-up, staged disclosure) \\
    - **Comparison** (contrast groups, time periods, categories, or benchmarks) \\
    - **Ellipsis** (de-emphasize less important data to focus attention) \\

    These strategies should be applied deliberately and coherently. \\

    ================================================== \\
    INPUT INFORMATION \\
    ================================================== \\

    Intent of the scene: \\
    {row['Intent']} \\

    Chart Data (Ground Truth): \\
    {row['Chart Data']} \\

    ================================================== \\
    STRUCTURE NOTES \\
    ================================================== \\

    - Number of charts in this sample: 
    \{chart\_count\} \\
    - Number of reference images: 
    \{image\_count\} \\

    ================================================== \\
    IMAGE STYLE GUIDANCE (IMPORTANT) \\
    ================================================== \\

    - Closely follow the visual style of the reference images: \\
      - color palette and color family \\
      - typography style and weight \\
      - background treatment \\
      - overall visual density and tone \\

    ================================================== \\
    ANIMATION GUIDANCE \\
    ================================================== \\

    - If multiple charts exist, animate them sequentially or jointly \\
    - Animation timing must support the narrative you create \\
    - The animation MUST still produce a correct static visualization if motion fails \\

    ================================================== \\
    OUTPUT FORMAT (STRICT) \\
    ================================================== \\

    - Output MUST be a single, complete HTML document \\
    - Start with <!DOCTYPE html> \\
    - Include <html>, <head>, and <body> tags \\
    - Include exactly ONE <svg id="chart" width="1280" height="720"> \\
    - Include all JavaScript inline inside <script> tags \\
    - Define and invoke scheduleNow() exactly once at the end \\
    - Do NOT include explanations, comments outside HTML, or markdown \\
    - Return ONLY the raw HTML text, nothing else \\

    \end{tcolorbox}

    \caption{Data Video Generator system prompt — subtitle constraints, narrative strategies, input structure, and output format.}
    \label{fig:datavideo_system_prompt_part2}
\end{figure*}

\begin{figure*}[t]
    \centering
    
    \begin{tcolorbox}[
        colback=gray!5!white,
        colframe=gray!60!black,
        title=\textbf{Director Prompt},
        fonttitle=\bfseries\small,
        boxrule=0.8pt,
        arc=2pt,
        left=2pt, right=2pt, top=2pt, bottom=2pt
    ]
    \ttfamily\scriptsize
    \sloppy
    \setlength{\emergencystretch}{3em}
    You are a chart animation director. \\
    \\
    The goal is to create an **animated visual story** that fulfills the stated \\
    **intent of the scene**, using ONLY the provided data. \\
    \\
    IMPORTANT: An image (ss.png) has been provided representing the required visual style. \\
    Analyze the image to plan the animation layout. \\
    \\
    ================================================== \\
    INPUT INFORMATION \\
    ================================================== \\
    \\
    Intent of the scene: \\
    \{intent\} \\
    \\
    Chart Data (Ground Truth): \\
    \{data\} \\
    \\
    Time limit: \{duration\} seconds \\
    \\
    ================================================== \\
    CORE REQUIREMENT (VERY IMPORTANT) \\
    ================================================== \\
    \\
    - The animation MUST focus on expressing the **intent** through visual storytelling. \\
    - Author a cohesive STORY that conveys the intent using animation. \\
    - The story must be constructed from the provided data ONLY. \\
    - The narrative must be expressed through **on-screen subtitles** and **visual animation**. \\
    - The animation + subtitles together must clearly fulfill the stated **intent**. \\
    - Subtitles should explain, highlight, compare, or summarize the data as needed. \\
    - The story should unfold progressively over time, not all at once. \\
    \\
    ================================================== \\
    NARRATIVE ANIMATION STRATEGIES (GUIDANCE) \\
    ================================================== \\
    \\
    Choose animation strategies that best support the **intent** and the story you \\
    are constructing from the data: \\
    \\
    - **Emphasis**: highlight key values, peaks, outliers, or categories \\
    - **Suspense**: gradual reveal, delayed comparison, count-up, staged disclosure \\
    - **Comparison**: contrast groups, time periods, categories, or benchmarks \\
    - **Ellipsis**: de-emphasize less important data to focus attention \\
    \\
    These strategies should be applied deliberately and coherently. \\
    \\
    ================================================== \\
    TASK \\
    ================================================== \\
    \\
    1. Produce a structured animation plan (JSON). \\
    2. The plan must specify how to replicate the layout, chart type, and positioning seen in the provided image. \\
    3. Create a "subtitles" array within the JSON. Each entry must have 'start', 'end', and 'text'. \\
    4. Ensure the visual animation stages align perfectly with these subtitle timestamps. \\
    \\
    Return JSON only. \\
    \end{tcolorbox}
    \caption{Director prompt used to produce a structured animation plan aligned with the scene intent and reference visual style.}
    \label{fig:director_prompt}
\end{figure*}

\begin{figure*}[t]
    \centering
    
    \begin{tcolorbox}[
        colback=gray!5!white,
        colframe=gray!60!black,
        title=\textbf{Plan Critic Prompt},
        fonttitle=\bfseries\small,
        boxrule=0.8pt,
        arc=2pt,
        left=2pt, right=2pt, top=2pt, bottom=2pt
    ]
    \ttfamily\scriptsize
    \sloppy
    \setlength{\emergencystretch}{3em}
    You are a senior animation consultant. Review the plan against the source data, intent, and the PROVIDED IMAGES for visual style. \\
    \\
    Your primary task is to verify that the plan **fulfills the stated intent** of the scene. \\
    \\
    ================================================== \\
    INPUT INFORMATION \\
    ================================================== \\
    \\
    Intent of the scene: \\
    \{intent\} \\
    \\
    Chart Data (Ground Truth): \\
    \{data\} \\
    \\
    Time limit: \{duration\} seconds \\
    \\
    ================================================== \\
    PROPOSED PLAN \\
    ================================================== \\
    \\
    \{plan\} \\
    \\
    ================================================== \\
    CORE REQUIREMENT (VERY IMPORTANT) \\
    ================================================== \\
    \\
    - The plan must author a cohesive STORY that conveys the intent using animation. \\
    - The story must be constructed from the provided data ONLY. \\
    - The narrative must be expressed through **on-screen subtitles** and **visual animation**. \\
    - The animation + subtitles together must clearly fulfill the stated **intent**. \\
    - Subtitles should explain, highlight, compare, or summarize the data as needed. \\
    - The story should unfold progressively over time, not all at once. \\
    \\
    ================================================== \\
    NARRATIVE ANIMATION STRATEGIES (GUIDANCE) \\
    ================================================== \\
    \\
    Ensure the plan uses appropriate animation strategies: \\
    \\
    - **Emphasis**: highlight key values, peaks, outliers, or categories \\
    - **Suspense**: gradual reveal, delayed comparison, count-up, staged disclosure \\
    - **Comparison**: contrast groups, time periods, categories, or benchmarks \\
    - **Ellipsis**: de-emphasize less important data to focus attention \\
    \\
    ================================================== \\
    CRITIQUE CHECKLIST \\
    ================================================== \\
    \\
    - **intent fulfillment**: Does the plan clearly express and fulfill the stated intent? Is the intent the central focus of the animation? \\
    - visual style: Does the plan's description align with the visual style provided in the images? \\
    - accuracy: Is the data representation correct? \\
    - timing: Is the flow realistic for \{duration\} seconds? \\
    - narrative: Does the plan tell a coherent story that supports the intent? \\
    - strategies: Are the animation strategies applied deliberately to reinforce the intent? \\
    \\
    Return concise, actionable feedback. If the intent is not fulfilled, explain what is missing and how to fix it. \\
    \end{tcolorbox}
    \caption{Plan Critic prompt used to validate intent fulfillment, data accuracy, style alignment, and timing of the proposed plan.}
    \label{fig:plan_critic_prompt}
\end{figure*}


\begin{figure*}[t]
    \centering
    
    \begin{tcolorbox}[
        colback=gray!5!white,
        colframe=gray!60!black,
        title=\textbf{Coder Prompt},
        fonttitle=\bfseries\small,
        boxrule=0.8pt,
        arc=2pt,
        left=2pt, right=2pt, top=2pt, bottom=2pt
    ]
    \ttfamily\scriptsize
    \sloppy
    \setlength{\emergencystretch}{3em}
    You are a D3.js animation engineer generating a **self-contained HTML file** that animates charts for a data video scene. \\
    \\
    IMPORTANT: Images have been provided representing the required visual style. \\
    You MUST analyze the attached images to extract and replicate: \\
    1. Exact Color Palette: Hex codes for background, marks, and text. \\
    2. Typography: Match font style and sizing. \\
    3. Layout: Replicate padding and positioning of elements. \\
    \\
    ================================================== \\
    STRICT RULES \\
    ================================================== \\
    \\
    - Output ONLY valid HTML \\
    - Use exactly ONE <svg id="chart"> \\
    - The SVG represents the FULL video frame \\
    - SVG MUST use a fixed resolution of 1280x720 pixels \\
    - <svg id="chart" width="1280" height="720"> \\
    - Do NOT resize the SVG dynamically \\
    \\
    - Define: \texttt{window.\_\_VIDEO\_DURATION\_\_} = \{duration\} \\
    - THIS TIMING IS STRICTLY ENFORCED DURING RENDERING \\
    - Use the entire duration meaningfully \\
    - Allow the story to unfold progressively \\
    - Avoid finishing too early or rushing key moments \\
    \\
    - Define functions: resetChart() and scheduleNow() \\
    - scheduleNow() MUST produce a complete animation every time it is called \\
    - resetChart() MUST restore the SVG to a valid initial state \\
    - Use setTimeout for ALL animations \\
    - NEVER reference requestAnimationFrame (directly or indirectly) \\
    - Do NOT load external assets except the D3 CDN \\
    - Do NOT invent, interpolate, or modify data values \\
    - Deterministic animation only (no randomness, no frame-based callbacks) \\
    \\
    ================================================== \\
    SUBTITLE REQUIREMENTS (MANDATORY) \\
    ================================================== \\
    \\
    - You MUST generate subtitles to narrate the story \\
    - Subtitles for a scene must be long enough to be readable by the viewer \\
    - Subtitles must appear within the SVG (not HTML overlays) \\
    - Subtitles must be synchronized with animation events \\
    - Subtitles must explain the story implied by the intent \\
    - Subtitles must be readable, non-overlapping, and stay within SVG bounds \\
    - Subtitles MUST fit entirely inside the 1280x720 frame at all times \\
    - Subtitle text must wrap or line-break to avoid overflow \\
    - Subtitles must never be clipped or cropped \\
    - Use a consistent subtitle position (e.g., bottom-center) \\
    \\
    ================================================== \\
    LAYOUT \& LEGIBILITY CONSTRAINTS \\
    ================================================== \\
    \\
    - All charts, subtitles, axes, labels, and annotations MUST fit within 1280x720 \\
    - Use explicit inner margins; do NOT place marks on SVG edges \\
    - Reserve space for subtitles and annotations \\
    - Ensure no overlap between visual elements \\
    - Axis labels, tick labels, annotations, and subtitles must not collide \\
    - Prioritize clarity and readability over visual flair \\
    \\
    ================================================== \\
    PLAN: \\
    ================================================== \\
    \\
    \{plan\} \\
    \\
    FEEDBACK TO ADDRESS: \\
    \{feedback\} \\
    \\

    \end{tcolorbox}
    \caption{Coder prompt used to generate a single self-contained D3 HTML animation that follows the plan and addresses critique feedback.}
    \label{fig:coder_prompt}
\end{figure*}

\begin{figure*}[t]
    \centering
    
    \begin{tcolorbox}[
        colback=gray!5!white,
        colframe=gray!60!black,
        title=\textbf{Video Critic Prompt},
        fonttitle=\bfseries\small,
        boxrule=0.8pt,
        arc=2pt,
        left=2pt, right=2pt, top=2pt, bottom=2pt
    ]
    \ttfamily\scriptsize
    \sloppy
    \setlength{\emergencystretch}{3em}
    Evaluate the rendered video against the source data, the plan, and the PROVIDED IMAGES. \\
    \\
    Your primary task is to verify that the video **expresses the stated intent** through its animations and subtitles. \\
    \\
    ================================================== \\
    INPUT INFORMATION \\
    ================================================== \\
    \\
    Intent of the scene: \\
    \{intent\} \\
    \\
    Chart Data (Ground Truth): \\
    \{data\} \\
    \\
    ================================================== \\
    CORE REQUIREMENT (VERY IMPORTANT) \\
    ================================================== \\
    \\
    - The video must tell a cohesive STORY that conveys the intent using animation. \\
    - The story must be constructed from the provided data ONLY. \\
    - The narrative must be expressed through **on-screen subtitles** and **visual animation**. \\
    - The animation + subtitles together must clearly fulfill the stated **intent**. \\
    - Subtitles should explain, highlight, compare, or summarize the data as needed. \\
    - The story should unfold progressively over time, not all at once. \\
    \\
    ================================================== \\
    ANIMATION ASSESSMENT (CRITICAL) \\
    ================================================== \\
    \\
    Carefully evaluate the animations in the video: \\
    \\
    1. **Animation Correctness**: Are the animations taking place accurately or are there issues like overlapping, clipping, or misplacement of text or visual elements? \\
    2. **Time Utilization**: Is the video using the entire allotted duration effectively to tell the story? it is too fast or too slow? Does it end too early or rush key moments? \\
    3. **Intent Expression**: Do the animations effectively express and support the intent? \\
    4. **Animation-Subtitle Sync**: Are animations properly synchronized with subtitles? Do they appear together at the right moments? \\
    5. **Animation Quality**: Are animations smooth, visible, and purposeful? Are there too many, too few, or poorly timed animations? \\
    6. **Animation Effectiveness**: Do the animations help the viewer understand the data story? \\
    \\
    Based on your assessment, provide specific feedback to: \\
    - **ADD** animations if key moments lack visual emphasis \\
    - **REMOVE** animations if they are distracting or redundant \\
    - **CHANGE** animations if timing, duration, or style needs adjustment \\
    - **EDIT** animations with subtitles if they are misaligned \\
    \\
    If perfect, start with 'PASS'. Otherwise, provide specific D3.js/CSS fixes with clear instructions on what animations to add, remove, change, or re-sync. \\
    \end{tcolorbox}
    \caption{Video Critic prompt used to evaluate the rendered animation for intent expression, correctness, pacing, and subtitle synchronization.}
    \label{fig:video_critic_prompt}
\end{figure*}

\begin{figure*}[t]
    \centering
    
    \begin{tcolorbox}[
        colback=gray!5!white,
        colframe=gray!60!black,
        title=\textbf{HTML Judge Prompt},
        fonttitle=\bfseries\small,
        boxrule=0.8pt,
        arc=2pt,
        left=2pt, right=2pt, top=2pt, bottom=2pt
    ]
    \ttfamily\scriptsize
    \sloppy
    \setlength{\emergencystretch}{3em}
    
    You are a critical evaluator assessing an automatically generated **data video**. \\
    \\
    SAMPLE ID: \{sample\_id\} \\
    \\
    You are given: \\
    - The complete HTML source used to render the visualization \\
    - The intent of the scene \\
    - The original narration transcript \\
    - The data used to generate the visualization \\
    \\
    IMPORTANT: \\
    - Base your judgment ONLY on the provided HTML and text \\
    - Do NOT assume animation quality beyond what the HTML explicitly implies \\
    - Be very strict and objective in your evaluation \\
    \\
    ================================================== \\
    INPUT DATA \\
    ================================================== \\
    \\
    Intent: \{row['Intent']\} \\
    Original Transcript: \{row['Transcript']\} \\
    Chart Data: \{row['Chart Data']\} \\
    HTML Source (FULL): \{html\} \\
    ================================================== \\
    EVALUATION CRITERIA \\
    ================================================== \\
    \\
    1. **Narrative Quality (0–5)** \\
    Evaluate whether the intended message of the scene is covered and conveyed as a coherent story through the sequence of animation scenes and subtitles. \\
    \\
    - 0: The animation does not reflect the stated intent and lacks any coherent story. \\
    - 1: The intent is barely addressed and the story is largely unclear or incomplete. \\
    - 2: The intent is partially covered, but the story is fragmented or poorly connected. \\
    - 3: The intent is mostly covered, but the narrative has noticeable gaps or weak transitions. \\
    - 4: The intent is clearly conveyed as a coherent story, with minor lapses in flow. \\
    - 5: The full intent is clearly and consistently conveyed through a well-structured narrative across all animation scenes and subtitles. \\
    \\
    2. **Informativeness (0–5)** \\
    Evaluate how insightful the subtitles and accompanying animations are, beyond simply reading or restating the data. \\
    \\
    - 0: Subtitles provide no meaningful information or insight. \\
    - 1: Subtitles primarily read out data values with little to no interpretation. \\
    - 2: Some interpretation is present, but subtitles remain mostly descriptive or obvious. \\
    - 3: Subtitles provide moderate insights, though some segments still rely on data restatement. \\
    - 4: Subtitles frequently highlight meaningful patterns, trends, or comparisons supported by animation. \\
    - 5: Subtitles consistently deliver clear, non-obvious insights into the data, effectively reinforced by animation. \\
    \\
    3. **Subtitle–Transcript Similarity (0–5)** \\
    Evaluate how well the generated subtitles align with the original narration transcript. \\
    \\
    - 0: Completely diverges from transcript intent. \\
    - 1: Largely contradicts transcript. \\
    - 2: Partial overlap with missing key ideas. \\
    - 3: General alignment with omissions. \\
    - 4: Strong alignment with minor differences. \\
    - 5: Preserves transcript meaning throughout. \\
    \\
    4. **Code Correctness (0–5)** \\
    Evaluate whether the **HTML code structure and logic correctly support the narration’s intended animation plan**. \\
    \\
    - 0: HTML code is clearly incorrect, broken, or unrelated to the intended narration. \\
    - 1: Major structural or logical issues prevent the HTML from supporting the narration. \\
    - 2: Some parts align with the narration, but key elements are missing or incorrect. \\
    - 3: HTML generally follows the narration plan, with noticeable inconsistencies or gaps. \\
    - 4: HTML correctly implements the narration plan with minor issues. \\
    - 5: HTML code cleanly and correctly implements the intended narration and animation plan throughout. \\
    \\

    \end{tcolorbox}
    
    \caption{Prompt for the HTML-based VLM judge used to evaluate generated data videos based on narrative quality, informativeness, subtitle–transcript alignment, and code correctness.}
    \label{fig:html_judge_prompt}
\end{figure*}

\begin{figure*}[t]
    \centering
    
    \begin{tcolorbox}[
        colback=gray!5!white,
        colframe=gray!60!black,
        title=\textbf{Pairwise VLM Judge Prompt},
        fonttitle=\bfseries\small,
        boxrule=0.8pt,
        arc=2pt,
        left=2pt, right=2pt, top=2pt, bottom=2pt
    ]
    \ttfamily\scriptsize
    \sloppy
    \setlength{\emergencystretch}{3em}
    
    You are a strict expert evaluator comparing two automatically generated **data videos**. \\
    \\
    SAMPLE ID: \{sample\_id\} \\
    \\
    You are given: \\
    1. **Video A** - A generated data video (MP4) \\
    2. **Video B** - Another generated data video (MP4) \\
    3. A **reference screenshot** showing the expected visual style \\
    \\
    Your task is to compare Video A and Video B based on the criteria below and determine which video is better overall. \\
    \\
    ================================================== \\
    EVALUATION CRITERIA \\
    ================================================== \\
    \\
    -------------------------------------------------- \\
    1. **Visualization Quality** \\
    \\
    Definition: \\
    Evaluate whether the visualization is rendered correctly and remains visually readable. \\
    \\
    Consider: \\
    - Are there rendering issues (blank screens, missing charts, or visual glitches)? \\
    - Is the chart readable (no overlapping elements, clipping, or unreadable text)? \\
    - Is the visualization clean, legible, and well-structured? \\
    \\
    -------------------------------------------------- \\
    2. **Subtitle-Animation Coherence** \\
    \\
    Definition: \\
    Evaluate the alignment between what the subtitles say and what the animation shows. \\
    \\
    Consider: \\
    - Do the subtitles match what is being shown in the animation? \\
    - Is there proper timing and synchronization? \\
    - Do the subtitles and visuals reinforce each other? \\
    \\
    -------------------------------------------------- \\
    3. **Style Consistency** \\
    \\
    Definition: \\
    Evaluate how well each video maintains the visual style shown in the reference screenshot. \\
    \\
    Consider: \\
    - Does the video use similar colors, chart type, layout, and design elements? \\
    - How closely does each video follow the reference style? \\
    \\
    ================================================== \\
    OUTPUT FORMAT (STRICT JSON ONLY) \\
    ================================================== \\
    \\
    Return ONLY valid JSON. \\
    Do NOT include extra text. \\
    \\
    Use EXACTLY this format: \\
    \\
    \{\{ \\
    \hspace*{1em}"visualization\_quality": \{\{ \\
    \hspace*{2em}"video\_a\_analysis": "1-3 sentences analyzing Video A", \\
    \hspace*{2em}"video\_b\_analysis": "1-3 sentences analyzing Video B", \\
    \hspace*{2em}"better\_video": "A" or "B" or "tie" \\
    \hspace*{1em}\}\}, \\
    \hspace*{1em}"subtitle\_animation\_coherence": \{\{ \\
    \hspace*{2em}"video\_a\_analysis": "1-3 sentences analyzing Video A", \\
    \hspace*{2em}"video\_b\_analysis": "1-3 sentences analyzing Video B", \\
    \hspace*{2em}"better\_video": "A" or "B" or "tie" \\
    \hspace*{1em}\}\}, \\
    \hspace*{1em}"style\_consistency": \{\{ \\
    \hspace*{2em}"video\_a\_analysis": "1-3 sentences analyzing Video A", \\
    \hspace*{2em}"video\_b\_analysis": "1-3 sentences analyzing Video B", \\
    \hspace*{2em}"better\_video": "A" or "B" or "tie" \\
    \hspace*{1em}\}\}, \\
    \hspace*{1em}"overall\_verdict": \{\{ \\
    \hspace*{2em}"rationale": "2-4 sentences explaining the overall comparison", \\
    \hspace*{2em}"winner": "A" or "B" or "tie" \\
    \hspace*{1em}\}\} \\
    \}\}
    
    \end{tcolorbox}
    
    \caption{Prompt for the pairwise VLM judge used to compare two generated data videos across visualization quality, subtitle--animation coherence, and style consistency.}
    \label{fig:pairwise_vlm_judge_prompt}
\end{figure*}

\end{appendices}